\documentclass[11pt]{article}

\usepackage[final]{acl}
\usepackage{soul, xcolor}
\usepackage{pifont}
\sethlcolor{yellow} 
\usepackage{times}
\usepackage{latexsym}
\usepackage{amsmath}
\usepackage{cleveref}
\usepackage{verbatim}
\usepackage[T1]{fontenc}

\usepackage[utf8]{inputenc}

\usepackage{microtype}
\DeclareUnicodeCharacter{2212}{\ensuremath{-}}

\usepackage{inconsolata}

\usepackage[most]{tcolorbox}
\usepackage{caption} 

\tcbset{
  promptbox/.style={
    colback=gray!10,
    colframe=gray!50,
    boxrule=0.6pt,
    arc=4pt,
    left=6pt,right=6pt,top=6pt,bottom=6pt,
    width=\textwidth,   
    boxsep=0pt,
  }
}

\usepackage{booktabs}
\usepackage{graphicx}
\usepackage{makecell}
\usepackage{multirow}
\usepackage{multicol}
\usepackage{enumitem}
\usepackage{amssymb}
\usepackage{colortbl}
\usepackage{tabularx}
\usepackage{array}

\title{\textsc{LLaTiSA}: Towards Difficulty-Stratified Time Series Reasoning from Visual Perception to Semantics}

\author{
  \textbf{Yueyang Ding\thanks{Work done at internship at Amap, Alibaba.}, HaoPeng Zhang, Rui Dai\thanks{Corresponding author.}, Yi Wang,} \\
  \textbf{Tianyu Zong, Kaikui Liu, Xiangxiang Chu} \\
  Amap, Alibaba Group \\
  \texttt{dyyyyy494@gmail.com, zongtianyu20@mails.ucas.ac.cn} \\
  \texttt{\{zhanghaopeng.zhp, daima.dr, zhiyi.wy, damon, chuxiangxiang.cxx\}@alibaba-inc.com}
}

\begin{document}
\maketitle
\begin{abstract}
Comprehensive understanding of time series remains a significant challenge for Large Language Models (LLMs). Current research is hindered by fragmented task definitions and benchmarks with inherent ambiguities, precluding rigorous evaluation and the development of unified Time Series Reasoning Models (TSRMs). To bridge this gap, we formalize Time Series Reasoning (TSR) via a four-level taxonomy of increasing cognitive complexity. We introduce \textsc{HiTSR}, a \textbf{hi}erarchical \textbf{t}ime \textbf{s}eries \textbf{r}easoning dataset comprising 83k samples with diverse task combinations and verified Chain-of-Thought (CoT) trajectories. Leveraging \textsc{HiTSR}, we propose \textsc{LLaTiSA}, a strong TSRM that integrates visualized patterns with precision-calibrated numerical tables to enhance the temporal perception of Vision-Language Models (VLMs). Through a multi-stage curriculum fine-tuning strategy, \textsc{LLaTiSA} achieves superior performance and exhibits robust out-of-distribution generalization across diverse TSR tasks and real-world scenarios. Our code is available at \href{https://github.com/RainingNovember/LLaTiSA}{https://github.com/RainingNovember/LLaTiSA}.
\end{abstract}

\section{Introduction}


Time series data serve as a fundamental modality for capturing the temporal dynamics of complex systems, underpinning a wide range of critical applications, including medical diagnostics \cite{heartcare}, financial valuation \cite{finmultitime}, and industrial condition monitoring \cite{itformer}. Interpreting real-world temporal dynamics necessitates Time Series Reasoning (TSR): deriving a comprehensive understanding of time series by grounding analysis in numerical evidence, perceiving salient patterns, and incorporating contextual information to support challenging tasks such as decision-making or forecasting. 
\begin{figure*}[t]
    \centering
    \includegraphics[width=0.85\linewidth]{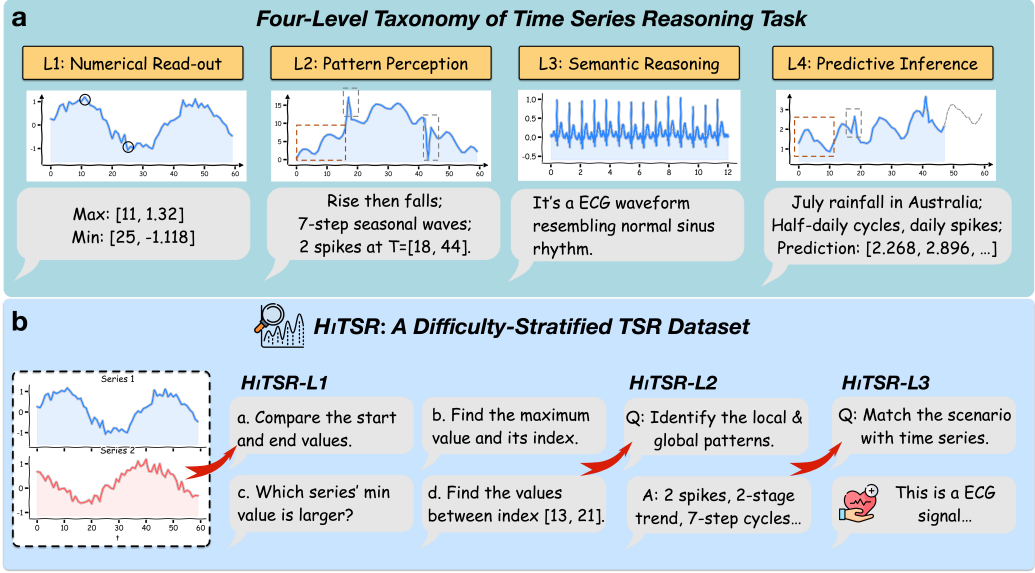}
    \caption{(a) Illustration of the four-level TSR taxonomy and \textsc{HiTSR} dataset. Each incorporates a simple TSR case with increasing difficulties. (b) A brief introduction to \textsc{HiTSR} datasets, containing three levels of TSR scenarios with diverse subtasks.}
    \label{fig:1}
\end{figure*}

Recent studies have demonstrated that Large Language Models (LLMs) can be effective for time-series understanding \cite{chatts} and provide interpretable reasoning to further enrich traditional time series analysis \cite{timemaster}. 
Despite these advancements, building unified TSRMs faces two critical bottlenecks: \textbf{(1) The lack of a formalized taxonomy for reasoning tasks.} Existing suites like TSQA \cite{timemqa} or BEDTime \cite{bedtime} either present a mapping of traditional analytical tasks (e.g., forecasting and classification) to text formats or focus on realizing scenario-agnostic series patterns. In contrast, domain-specific datasets like Engine-QA \cite{itformer} and Heartcare-Bench \cite{heartcare} tend to be numerically-insensitive, primarily relying on the semantic relationships between multiple series. \textbf{(2) Reliability deficits in existing benchmarks.} Beyond insufficient contextual information, current benchmarks are often hampered by semantic ambiguities in label definitions and low-fidelity question-answer pairs \cite{timeomni}. To facilitate the systematic development and rigorous assessment of TSRMs, it is essential to formalize unified definitions of TSR and construct reliable benchmarks that comprehensively model capabilities across diverse scenarios.

Building on the aforementioned limitations, we posit that cultivating reliable TSR capabilities mirrors a multi-stage cognitive process: transitioning from point-level numerical grounding to series-level perception, facilitating high-level semantic interpretation, and ultimately enabling context-aware generation. 
Guided by this philosophy, we formalize a systematic taxonomy that decomposes TSR into four levels of increasing complexity (L1-L4), as shown in \Cref{fig:1}(a). Leveraging this taxonomy, we introduce \textsc{HiTSR} (\Cref{fig:1}(b)), a \textbf{hi}erarchical \textbf{t}ime \textbf{s}eries \textbf{r}easoning dataset comprising 83k samples that span the foundational TSR tasks across the first three levels (L1-L3). All samples in \textsc{HiTSR} are constructed with unambiguous ground truths and verifiable reasoning chains. 
\textbf{This paper focuses on levels L1-L3, since we find that even advanced TSRMs struggle significantly on these foundational tasks}\footnote{We analyze these findings in \Cref{subsec:model_findings,sec:eval_hits}.}, suggesting that mastering these fundamental stages is a prerequisite for reliable time-series reasoning and remains an immediate challenge.

Recent works have shown that Vision Language Models (VLMs) can excel in basic TSR tasks by relying exclusively on time series visualizations \cite{timemaster,bedtime}. However, such vision-centric approaches frequently struggle with numerical reasoning tasks that require precise value-based evidence. Inspired by \cite{deepseekocr}, which utilizes visual tokens to represent textual information efficiently, we introduce a robust VLM-based TSRM named \textbf{L}arge \textbf{L}anguage \textbf{a}nd \textbf{Ti}me \textbf{S}eries \textbf{A}ssistant (\textsc{LLaTiSA}). \textsc{LLaTiSA} is designed to bridge the gap between qualitative visual intuition and quantitative numerical precision. By employing a three-stage curriculum fine-tuning strategy aligned with our proposed L1-L3 hierarchy, \textsc{LLaTiSA} demonstrates that TSR benefits from incremental cognitive grounding and high-fidelity Chains-of-Thought (CoT). In summary, our contributions are summarized as follows:
\begin{itemize}[left=0pt]
\setlength{\itemsep}{0.5pt}
    \item \textbf{A cognitive taxonomy and unified dataset.} We formalize a four-level cognitive taxonomy (L1-L4) that captures the progressive nature of temporal understanding. Based on this, we introduce \textsc{HiTSR}, a large-scale, high-fidelity TSR dataset of over 83k samples spanning L1-L3, featuring diverse task settings and verified reasoning chains.
    \item \textbf{A numerically-enhanced TSRM.} We propose \textsc{LLaTiSA}, a VLM-based TSRM that bridges qualitative perception and quantitative precision. \textsc{LLaTiSA} is trained via a three-stage curriculum aligned with our L1-L3 hierarchy to build robust reasoning capabilities progressively.
    \item \textbf{Empirical validation of multi-stage reasoning.} Extensive experiments demonstrate that \textsc{LLaTiSA} consistently outperforms proprietary models and various open-source baselines. Furthermore, our model exhibits superior generalization to out-of-distribution (OOD) scenarios across existing benchmarks.
\end{itemize}

\section{Related Works}
\subsection{Time Series Reasoning Datasets}
Recently released multimodal time series datasets adopt various types of TSR tasks. For general question-answering, early efforts like TSQA \cite{timemqa} and TRQA \cite{trqa} explore LLMs' potential in question-answering and traditional time series analysis tasks. SciTS \cite{scits} introduces a benchmark specifically designed for scientific time series analysis across a wide range of scientific domains. RATs40k \cite{timera} transforms classical time series anomaly detection from discriminative into a reasoning-intensive task and annotates samples with explanatory reasoning. TimeMMD \cite{timemmd}is tailored for the context-aware time series forecasting task. MMTS-Bench \cite{mmtsbench} provides a multi-dimensional task classification taxonomy, yet it overlooks evaluating models' basic numerical read-out and combining diverse semantics of real scenarios. In addition to insufficient context, current benchmarks are further hampered by semantic ambiguities in label definitions and low-fidelity question-answer pairs \cite{timeomni}. Moreover, most of the datasets focus on either pattern description or complex reasoning scenarios such as forecasting, and overlook evaluating models' basic ability of read-out numerical values. Our findings suggest that current models even struggle with finding point-level numerical evidence, suffering from overthinking or hallucinations with longer time horizons. 

\subsection{Time Series Reasoning Models}
Understanding time series via language models generally mostly follows three encoding paradigms: (i) direct serialization into text tokens, (ii) visual rendering as plots for VLMs, and (iii) the integration of dedicated time-series encoders to construct Time-Series Multimodal Large Language Models (TS-MLLMs). TimeOmni-1 \cite{timeomni} and Time-R1 \cite{timer1} employ textual inputs for context-aware forecasting. ChatTS \cite{chatts} and ITFormer \cite{itformer} incorporate an MLP-based and a Q-former \cite{blip} style time series encoder to perform multivariate TSR tasks, respectively. TimeMaster \cite{timemaster} enables VLMs to perform interpretable classification based on visualized time series. While recent studies point out that TSRMs struggle with basic TSR tasks like description differentiation \cite{bedtime}, VLMs have demonstrated superior robustness compared to their LLM and TS-MLLM counterparts. This observation motivates us to further unlock the potential of VLMs for general-purpose TSR scenarios.

\section{Method}
\subsection{Time Series Reasoning Taxonomy}
We argue that reliable TSR is inherently compositional, necessitating a hierarchy that can: (i) faithfully ground observations to timestamps, (ii) abstract salient temporal patterns beyond point-level evidence, and (iii) integrate with contextual knowledge to support semantic judgment and prediction. 
In contrast, prior TSR benchmarks are typically organized around isolated tasks and adopt heterogeneous definitions of what constitutes ``reasoning'', often over-emphasizing a narrow subset of problem types (e.g., coarse pattern discrimination). 
Such fragmentation leads to two limitations. First, cross-benchmark comparability is compromised because mismatched capability sets and inconsistent task difficulties prevent a unified assessment of progress. Second, current TSRMs often attempt to perform high-level reasoning (e.g., forecasting) on an unsteady foundation: models still struggle with elementary grounding and verification, which undermines the credibility and interpretability of more complex reasoning. 

To address these limitations, we introduce a difficulty-stratified taxonomy that organizes TSR into progressively increasing levels of complexity. This framework serves as a unified schema for categorizing diverse TSR tasks and provides a diagnostic lens to pinpoint the specific cognitive boundaries of current models. Concretely, we decompose TSR into four hierarchical levels:
\begin{itemize}[left=0pt]
\setlength{\itemsep}{0.5pt}
    \item \textbf{L1:} \emph{Numerical Read-out}. Establish time-aware indexing and point-level numerical retrieval. 
    \item \textbf{L2:} \emph{Pattern Perception}. Identify and differentiate multi-scale temporal patterns using quantitative evidence.
    \item \textbf{L3:} \emph{Semantic Reasoning}. Integrate time series observations with contextual knowledge to perform domain-specific reasoning.
    \item \textbf{L4:} \emph{Predictive Inference}. Generate high-fidelity time-series predictions.
\end{itemize}
Based on such a taxonomy, we summarize current benchmarks into different levels in \Cref{sec:task_sum}, posing a general standard to task definition and evaluation of TSRMs' capabilities.

\begin{figure*}[t]
    \centering
    \includegraphics[width=0.85\textwidth]{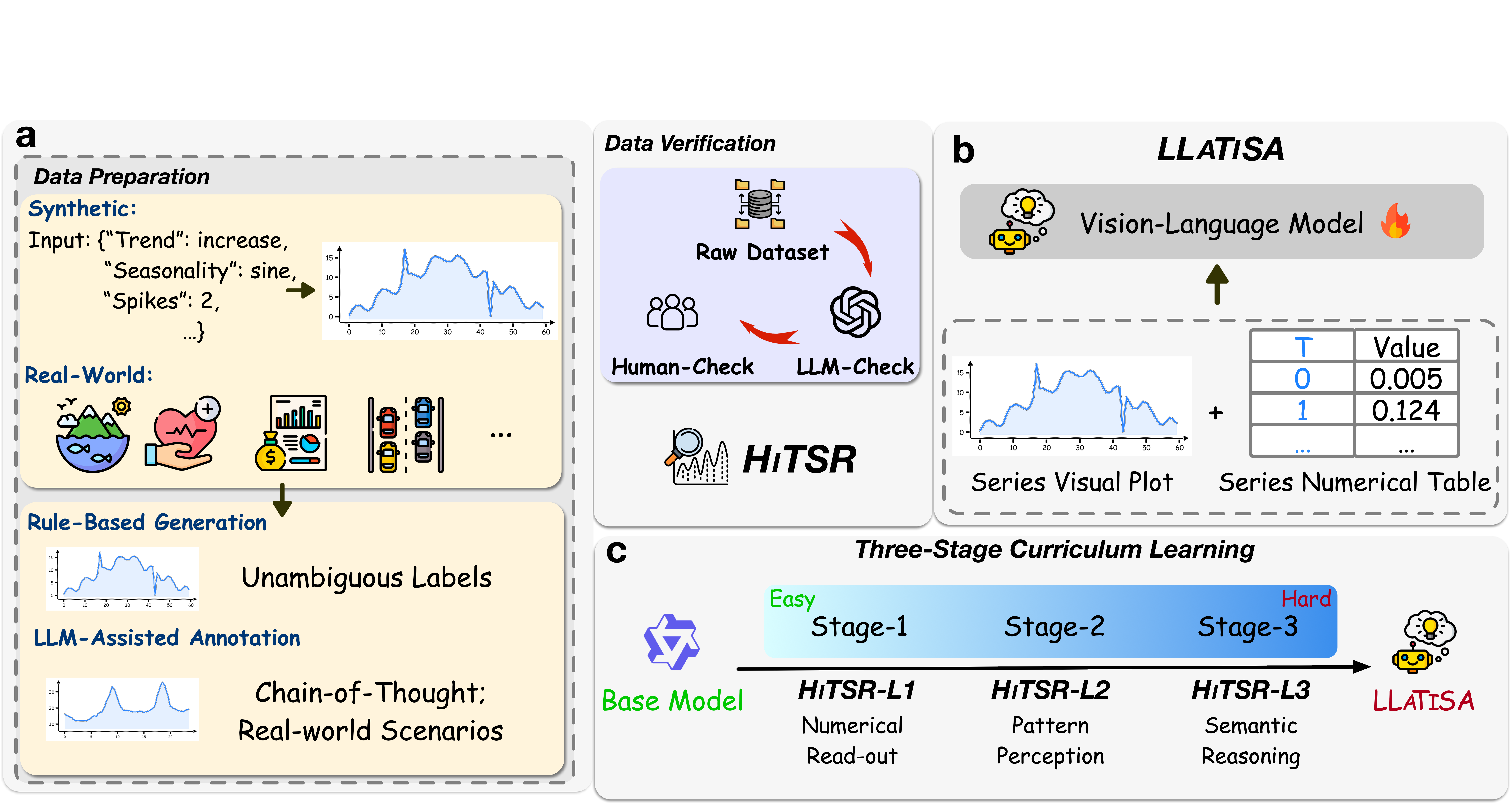}
    \caption{Overview of data pipeline and the model framework. (a) Constructions of \textsc{HiTSR}, including data collection, sample annotation, and multi-stage verification. (b) Framework of \textsc{LLaTiSA}, which incorporates dual-image inputs and performed (c) three-stage curriculum fine-tuning on the \textsc{HiTSR}.}
    \label{fig:2}
\end{figure*}

\subsection{\textsc{HiTSR} Dataset}
Building on our four-level TSR taxonomy, we introduce \textsc{HiTSR}, a unified dataset for both training and evaluation of TSR capabilities across levels L1-L3. Unlike prior benchmarks that may contain ambiguous queries or underspecified context, \textsc{HiTSR} dataset is constructed with rigorous validity checks to ensure exact problem statements and sufficient contextual information. Overall, \textsc{HiTSR} comprises approximately 83k samples spanning synthetic and real-world time series, annotated and verified via a multi-stage process.

\paragraph{Data Sources.}
In total, the \textsc{HiTSR} dataset comprises three hierarchical levels: \textsc{HiTSR-L1} ($30{,}000$ samples), \textsc{HiTSR-L2} ($50{,}703$ samples), and \textsc{HiTSR-L3} ($3{,}121$ samples). 
Each subset includes a high-fidelity test set rigorously verified by human annotators. 
For \textsc{HiTSR-L1} and \textsc{HiTSR-L2}, our primary objective is to provide large-scale supervision that strengthens fundamental reasoning abilities in a controlled yet diverse manner. 
We therefore generate \textsc{HiTSR-L1} and \textsc{HiTSR-L2} instances using synthetic time series sourced from \cite{chatts}. This approach allows us to systematically manipulate temporal structures and underlying patterns, while scaling the data to accommodate various training-budget requirements without being restricted by the scarcity of real-world annotations. For \textsc{HiTSR-L3}, we curate publicly available real-world time series from diverse domains and construct question-answer pairs written by the LLM analyzer. 
The complete data synthesis pipeline and the summary of real-world data information are detailed in \Cref{sec:data_source}.

\paragraph{Task Formulation.}
Following the proposed taxonomy, we instantiate a set of subtasks within each level to cover complementary facets of the TSR as shown in \Cref{sec:tasksum}. L1 tasks are formulated as short-answer questions, given that the ground truth can be generated via rule-based methods. Tasks within L2-L3 are structured as multiple-choice (MC) questions to ensure objective and reproducible evaluation. To maintain a strict instruction-following orientation and minimize biases from formatting artifacts, each supervision target is crafted as a complete natural-language statement rather than a concise label. Detailed task definitions and additional examples are provided in \Cref{sec:task_intro}. 
Additionally, we adopt different answer-construction protocols tailored to the degree of output determinacy. For L1 and the \emph{Numerical Perception} subtask in L2, the correct option can be derived by explicit rules from the underlying series; we therefore generate options programmatically under fixed templates. For the two \emph{Pattern Differentiation} subtasks in L2 and the L3 \emph{Semantic Understanding} task, the outputs are descriptive and require higher-level interpretation. We construct these items with an LLM-assisted pipeline and subsequently verify by humans to ensure correctness and remove ambiguous cases.

\paragraph{Multi-Stage Verified CoT Annotation.}
We adopt distinct annotation and verification pipelines for L2 and L3 tasks. For L2 \emph{Pattern Differentiation}, we first prompt GPT-5 with each time series’ visual plot and key statistics (e.g., extreme value-index pairs, spike locations with precise values, and notable local events) to produce numerically aware descriptions of local and global patterns. We then construct multiple-choice questions by sampling distractor options from descriptions of other series. 
To ensure reliability, we employ GPT-5 to cross-validate each question, removing ambiguous descriptions and flawed items (e.g., those with non-unique correct options). 
Finally, we hold out 10\% of the validated pool for evaluation. This subset undergoes a comprehensive human auditing, and only items that pass this rigorous review are retained, yielding a unambiguous test set. 
For L3 \emph{Semantic Understanding}, we leverage LLMs to generate scenario-aware multiple-choice questions by prompting them with real-world time series and their associated metadata. Each scenario description integrates pattern recognition and numerical grounding, effectively encapsulating the TSRMs' capabilities across L1-L3. 
Each item is subsequently dual-checked by GPT-5 and then reviewed by humans to comprehensively verify that the correct answer is supported by consistent numerical, pattern, and semantic evidence.

\subsection{\textsc{LLaTiSA}}
Time series can be presented to language models in various formats \cite{bedtime}. While purely visual plots effectively convey global shapes and local morphological cues, they often lack the fine-grained numerical precision required for grounded reasoning. Consequently, many TSR methods typically represent time series as textualized numerical sequences \cite{timeomni,timemqa} or integrate LLMs with dedicated time series encoders to develop TS-MLLMs \cite{chatts,timesense}. Recent studies, however, indicate that TS-MLLMs frequently underperform relative to general-purpose LLMs or VLMs on L2 TSR tasks, while VLMs are stronger on pattern-centric tasks (e.g., local pattern differentiation) \cite{bedtime}, underscoring the advantages of visual representations for L2 reasoning. 

To empower VLMs with precise numerical grounding, we propose \textsc{LLaTiSA} (\Cref{fig:2}.b), a dual-view input framework that pairs standard time series visualizations with a secondary image rendering the data as a structured index-value table. This design facilitates a progressive interpretation paradigm: the first plot enables macroscopic inspection for overall perception, the image of the structured table enables point-accurate verification of specific indices or intervals. By synthesizing these dual modalities alongside textual context, the model produces semantically grounded reasoning. Empirically, compared to processing series as textual arrays, \textsc{LLaTiSA} provides extra index information, thereby significantly mitigating numerical hallucinations and improving performance on numerical-sensitive tasks.

Leveraging \textsc{HiTSR}, \textsc{LLaTiSA} is trained via a three-stage curriculum designed to progressively acquire the prerequisite capabilities defined in our taxonomy.
During the initial two stages, we perform sequential Supervised Fine-Tuning (SFT) on \textsc{HiTSR-L1} and \textsc{HiTSR-L2} to consolidate the model's numerical read-out precision and pattern perception capabilities. 
In the third stage, \textsc{LLaTiSA} could be either trained on \textsc{HiTSR-L3} or general-purpose domain-specific TSR benchmarks. Our experiments further show that existing TSRMs often fail even on basic L1-L2 TSR, and struggle to combine LLM priors with series-level evidence for reliable L3 semantic inference. In contrast, curriculum-trained \textsc{LLaTiSA} across L1-L2 could be easily transferred to domain-specific L3 tasks (e.g., ECG interpretation).
\begin{table*}[tbp]
\centering
\setlength{\tabcolsep}{8pt}
\renewcommand{\arraystretch}{0.9}
\resizebox{\textwidth}{!}{
\begin{tabular}{ll|ccc|cc|c}
\toprule
\multirow{4}{*}{\textbf{Modality}} & \multirow{4}{*}{\textbf{Model}} & \multicolumn{3}{c|}{\textbf{L1 }\emph{Numerical Read-out}} & \multicolumn{2}{c|}{\textbf{L2 }\emph{Pattern Perception}} & \textbf{L3 }\emph{Semantic Reasoning} \\
\cmidrule(lr){3-8}
& & \multicolumn{3}{c|}{\textbf{Min \& Max Localization}} & \textbf{Local} & \textbf{Global} & \textbf{Series Comparison} \\
\cmidrule(lr){3-8}
& & Acc\% & half-Acc\% & SR\% & Acc\% & Acc\% & Acc\% \\
\midrule
\multirow{4}{*}{Text (w/o index)} 
 & GPT-4o & 47.4 & 69.8 & 100.0 & 47.6 & 95.0 & 43.0 \\
 & Qwen3-8B & 36.4 & 52.6 & 68.0 & 35.6 & 77.5 & 36.0 \\
 & Time-R1 & 0.0 & 16.0 & 100.0 & 32.4 & 46.7 & 43.0 \\
 & Time-MQA & 0.0 & 0.6 & 17.4 & 19.0 & 28.3 & 10.0 \\
\midrule
\multirow{2}{*}{Vision (plot)} 
 & GPT-4o & - & - & - & \underline{72.2} & \underline{96.7} & 39.0 \\
 & Qwen3-VL-8B & - & - & - & 38.2 & 85.8 & 41.0 \\
\midrule
\multirow{2}{*}{Vision (plot + num)} 
 & GPT-4o & 2.4 & 11.0 & 100.0 & 65.4 & 95.0 & 47.0 \\
 & Qwen3-VL-8B & 2.0 & 17.4 & 100.0 & 36.4 & 86.7 & 47.0 \\
\midrule
\multirow{2}{*}{Time Series} 
 & ChatTS & 7.8 & 29.2 & 100.0 & 57.0 & 80.0 & \underline{59.0} \\
 & OpenTSLM & - & - & - & 7.8 & 15.0 & 13.0 \\
\midrule
\multirow{2}{*}{Vision + Text (w/o index)} 
 & GPT-4o & 17.4 & 49.8 & 100.0 & 65.8 & 94.2 & 49.0 \\
 & Qwen3-VL-8B & 12.4 & 42.6 & 72.4 & 42.8 & 92.5 & 52.0 \\
\midrule
\multirow{2}{*}{Vision + Text (w/ index)} 
 & GPT-4o & \underline{54.2} & \underline{88.6} & 100.0 & 65.8 & \underline{96.7} & 48.0 \\
 & Qwen3-VL-8B & 34.2 & 68.0 & 90.6 & 43.0 & 93.3 & 54.0 \\
\midrule
Vision (plot + num) & \textsc{LLaTiSA} & \textbf{86.8} & \textbf{96.0} & 100.0 & \textbf{75.6} & \textbf{97.5} & \textbf{67.0} \\
\bottomrule
\end{tabular}
}
\caption{Performance comparison on \textbf{out-of-distribution datasets} (primarily prior benchmarks BEDTime, MMTS-Bench, and MCQ2, which differ substantially from the training data) across L1-L3. All tasks report accuracy (Acc), ``half-Acc'' represents the performance in locating one of the maximum or minimum values, and ``SR'' denotes whether the model provides valid answers or correctly maps target values with correct indices. For VLMs, ``plot'' and ``num'' denote rendering time series as a sequence plot and a numerical table, respectively. ``-'' denotes that the input modality cannot handle the corresponding task. The best result is \textbf{bolded} and the second best is in \underline{underlined}.}
\label{tab:zeroshot}
\end{table*}

\section{Experiment}
Our experiments focus on investigating the following research questions (RQs):
\begin{itemize}[left=0pt]
\setlength{\itemsep}{0.5pt}
\item\textbf{RQ1:} How does \textsc{LLaTiSA} perform in existing TSR benchmarks across levels L1-L3?
\item\textbf{RQ2:} How does \textsc{LLaTiSA} perform compared with different encoding strategies?
\item\textbf{RQ3:} How does \textsc{LLaTiSA} perform in generalizing to real-world applications?
\item\textbf{RQ4:} How does the curriculum tuning strategy and CoT-aware training data benefit \textsc{LLaTiSA}'s performance?
\end{itemize}
\subsection{Experiment Settings}
\label{sec:exp_settings}
To leverage the VLM's fundamental capability in interpreting numeric-centric images, \textsc{LLaTiSA} employs Qwen3-VL-8B-Instruct \cite{Qwen3-VL} as its backbone. 
To ensure a fair comparison, we report results exclusively on out-of-distribution (OOD) datasets across levels L1-L3 (see \Cref{tab:zeroshot}). To demonstrate that \textsc{LLaTiSA} serves as a strong foundation model when fine-tuned for a new domain, we further evaluate it on ECG interpretation under both in-distribution (ID) and OOD settings using the ECG-Grounding dataset \cite{ecgrounding}, as shown in \Cref{tab:ecg}.
Additionally, we conduct a hierarchical evaluation of baseline models on \textsc{HiTSR} and present findings regarding the limitations of current TSRMs (see \Cref{sec:eval_hits} for details). 
Comprehensive training specifications, including parameter configurations and the proportions of training datasets across levels L1-L3, are also provided in \Cref{sec:settings}.
\begin{table*}[tbp]
\centering
\setlength{\tabcolsep}{4pt}
\resizebox{\linewidth}{!}{
\renewcommand{\arraystretch}{1.05}
\begin{tabular}{c|c|cc|cc|c}
\toprule
\multirow{4}{*}{\textbf{Modality}} & \multirow{4}{*}{\textbf{Model}} & \multicolumn{2}{c|}{\textbf{L1 }\emph{Numerical Read-out}} & \multicolumn{2}{c|}{\textbf{L2 }\emph{Pattern Perception}} & \textbf{L3 }\emph{Semantic Reasoning} \\
\cmidrule(lr){3-7}
& & \multicolumn{2}{c|}{\textbf{Min \& Max Localization}} & \textbf{Local} & \textbf{Global} & \textbf{Series Comparison} \\
\cmidrule(lr){3-7}
& & Acc\% & SR\% & Acc\% & Acc\% & Acc\% \\
\midrule
\multirow{2}{*}{Text}
& w/ index & \textbf{87.6} & 100.0 & 39.0 & 94.2 & 47.0 \\
& w/o index & 65.2 & 92.4 & 37.6 & 89.2 & 43.0 \\
\midrule
\multirow{2}{*}{Vision}
 & Single image (plot) & - & - & \underline{68.0} & \textbf{97.5} & \underline{62.0} \\
 & Single image (num) & \underline{86.8} & 100.0 & 32.4 & 75.0 & 32.0 \\
\midrule
\multirow{2}{*}{\makecell{Multi- \\ Modalities}}
 & Plot + Text (w/ index) & 85.2 & 100.0 & 49.8 & \underline{95.8} & 60.0 \\
 & Plot + Text (w/o index) & 57.2 & 86.4 & 50.4 & 94.2 & 57.0 \\
\midrule
\multicolumn{1}{c|}{Vision}
 & \textsc{LLaTiSA} & \underline{86.8} & 100.0 & \textbf{75.6} & \textbf{97.5} & \textbf{67.0}\\
\bottomrule
\end{tabular}
}
\caption{Encoding strategy comparison in \textbf{out-of-distribution datasets} across L1-L3. ``Modality'' denotes the encoding format of time series. ``-'' denotes that the input modality cannot handle the corresponding task. All models follow the same training settings based on the Qwen3-8B or Qwen3-VL-8B backbones. All metrics are reported as averages in all subtasks across L1-L3.}
\label{tab:tsvlm}
\end{table*}

\paragraph{Baselines.} We conduct extensive experiments across various time series encoding strategies, including: (i) encoding series as textual arrays; (ii) visualizing as sequence plots or numerical grids (denoted as ``plot'' and ``num'' in our tables); (iii) adopting a dual-modality input comprising both visualizations and textual sequences; and (iv) leveraging TS-MLLMs.
Since \textsc{LLaTiSA} incorporates index information in the second image, we also construct two types of textual encoding methods, ``w/ index'' and ``w/o index'', to ensure fair comparison.
For closed-source models, we select GPT-4o \cite{gpt4}, utilizing it as both LLMs and VLMs. For open-source models, we employ Qwen3-8B \cite{qwen3} and its multimodal counterpart, Qwen3-VL-8B \cite{Qwen3-VL}, to evaluate the efficacy of our curriculum fine-tuning strategy and to conduct a comparative analysis of various training paradigms.
We also incorporate textual-encoding TSRMs, such as Time-R1 \cite{timer1} and Time-MQA \cite{timemqa}. 
For TS-MLLMs, we evaluate ChatTS \cite{chatts} alongside OpenTSLM \cite{opentslm}. For ECG interpretation, we include GEM \cite{ecgrounding} and PULSE \cite{pulse} as domain-specific methods. We adopt standardized instruction formats for different time series encoding strategies to ensure a fair comparison. Detailed information on baseline models and corresponding instruction templates is provided in \Cref{sec:baseline}.
\paragraph{Datasets.} 
For the L1 extreme value localization task, we construct OOD evaluations using the real-world subset of \textsc{HiTSR-L1}, given that \textsc{LLaTiSA} is trained exclusively on its synthetic counterpart.
Regarding L2, we employ BEDTime \cite{bedtime} and MMTS-Bench, which focus on local and global pattern differentiation, respectively. To balance inference overhead, we randomly sample 500 questions from BEDTime for testing, while the complete corresponding subset of MMTS-Bench (120 samples) is utilized.
For L3, we use MCQ2 \cite{mcq2} to assess multi-series contextual reasoning. 
Following the experimental setting of \cite{chatts}, we randomly sample 100 questions from MCQ2.
To further verify \textsc{LLaTiSA}'s potential as a robust foundation model for L3-level real-world multivariate TSR, we conduct domain-specific fine-tuning on ECG-Grounding \cite{ecgrounding}, assessing the model’s transferability across both ID and OOD scenarios.
\paragraph{Evaluation metrics.} We report accuracy (Acc) as the primary metric for evaluation across all tasks. Additionally, we report the success rate (SR) to assess the model's ability to generate valid answers.

\subsection{OOD Experiment Results}
As shown in \Cref{tab:zeroshot}, we conduct comprehensive OOD evaluations comparing \textsc{LLaTiSA} with various baselines across the L1-L3 levels. The results demonstrate that \textsc{LLaTiSA} achieves significant improvements across diverse TSR scenarios. Specifically, in the L1 point-level value localization task, we observe a markedly lower success rate on text-based models. These models frequently struggle to output valid answers, often becoming trapped in repetitive generation loops that lead to sequences exceeding the maximum context length.

\subsection{Impact of TS Representation Strategies}
To empirically validate the efficacy and efficiency of \textsc{LLaTiSA}'s dual-view encoding strategy, we perform a comprehensive comparison against various time-series representation strategies (introduced in \Cref{sec:exp_settings}) under the same backbone and training configuration, and evaluate performance on OOD benchmarks at levels L1-L3.
\paragraph{Performance Comparison.}
\Cref{tab:tsvlm} presents the comprehensive performance on OOD testbeds across various time series representation strategies. 
\textsc{LLaTiSA} consistently outperforms alternative strategies on the majority of OOD tasks across levels L1-L3, demonstrating its superior efficacy in TSR.
Under the curriculum learning strategy, almost all strategies show performance gains over their untuned counterparts. For text-based paradigms (notably the ``w/o index'' variant), the success rate for L1 tasks improves markedly with fine-tuning. Furthermore, incorporating explicit index information significantly improves the precision of point-level localization, leading to substantial improvements in both accuracy and SR.
\subsection{Application in ECG Interpretation}
To evaluate the generalization capabilities of \textsc{LLaTiSA} (previously fine-tuned on \textsc{HiTSR}), we further perform Supervised Fine-Tuning (SFT) on the ECG-Grounding 30k dataset. For a rigorous comparison, we also fine-tune Qwen3-VL-8B on the same dataset. Baseline results for ECG-specific models are directly sourced from \cite{ecgrounding}. The full results are shown in \Cref{tab:full_ecg}.

As illustrated in \Cref{tab:ecg}, \textsc{LLaTiSA} exhibits a distinct advantage in lead-wise evaluation, which directly reflects its adherence to the structured, 12-lead diagnostic procedure employed by professional clinicians. Specifically, \textsc{LLaTiSA} achieves remarkable gains in lead assessment coverage and accuracy, outperforming GEM (LLaVA) by 18.14\% and 14.22\% in the ID evaluation, respectively. These results indicate that \textsc{LLaTiSA} more consistently executes per-lead analysis and generates more reliable lead-specific statements. 
Compared to Qwen3-VL-8B, \textsc{LLaTiSA} demonstrates superior performance across all metrics, with notable improvements on diagnostic accuracy (5.42\%) and evidence-based reasoning (4.39\%) in the OOD scenario. 
Notably, \textsc{LLaTiSA} is fine-tuned on only 2.5\% of GEM’s instruction data (30k out of 1.186M samples). This high data efficiency demonstrates that \textsc{LLaTiSA} can be efficiently specialized for grounded ECG interpretation and highlights its potential as a general TSR foundation model for real-world applications.
\begin{table}[tbp]
\centering
\setlength{\tabcolsep}{1pt}
\renewcommand{\arraystretch}{1.2}
\resizebox{\linewidth}{!}{
\begin{tabular}{c|c|c|c|c|c}
\toprule
\textbf{Type} & \textbf{Model} & \makecell{\textbf{Diag. Acc.}} & \makecell{\textbf{L. Cov.}} & \makecell{\textbf{L. Acc.}} & \makecell{\textbf{Evi. Reas.}} \\
\midrule
\multirow{5}{*}{ID}  & PULSE         & 81.1 & 7.1 & 3.0 & 52.4 \\
                     & GEM (LLaVA)    & \textbf{87.2} & \underline{71.1} & \underline{46.4} & \textbf{75.1} \\
                     & GEM (PULSE)    & \underline{86.5} & 69.8 & 45.3 & \underline{74.7} \\
                     & Qwen3-VL-8B    & 60.9 & 69.3 & 50.1 & 63.8 \\
                     \cmidrule(lr){2-6}
                     & \textsc{LLaTiSA} & 62.8 & \textbf{84.0}& \textbf{53.0} & 71.2 \\
\midrule
\multirow{5}{*}{OOD} & PULSE         & 59.2 & 11.2 & 6.3 & 55.5 \\
                     & GEM (LLaVA)    & \underline{73.5} & \textbf{80.0} & \underline{49.0} & \underline{74.6} \\
                     & GEM (PULSE)    & \textbf{73.6} & \underline{78.9} & 48.0 & \textbf{75.4} \\
                     & Qwen3-VL-8B    & 59.0 & 56.4 & 38.1 & 63.8 \\
                     \cmidrule(lr){2-6}
                     & \textsc{LLaTiSA} & 62.2 & 66.5 & \textbf{49.2} & 66.6 \\
\bottomrule
\end{tabular}
}
\caption{Experimental results on the L3 ECG-Grounding task. Diag. Acc. stands for Diagnosis Accuracy; L. Cov. and L. Acc. denote Lead Assessment Coverage and Accuracy, respectively; Evi. Reas. represents Evidence-Based Reasoning. .}
\label{tab:ecg}
\end{table}
\subsection{Ablation Study}\label{sec:ablation}

\paragraph{Effectiveness of CoT.}
In the \textsc{HiTSR} dataset, we incorporate LLM-annotated CoT data, verified by humans, to enhance \textsc{LLaTiSA}'s performance and generalization, particularly on the more complex tasks at levels L2-L3. To isolate the impact of CoT, we perform SFT on the same dataset using only the correct options as supervision (``w/o CoT''). As shown in \Cref{tab:ablation}, while ID performance does not degrade significantly, the OOD performance deteriorates markedly, with accuracy decreasing by 4.31\% and 17.91\% on L2 and L3, respectively. 
Furthermore, training solely on correct options significantly undermines the model's original instruction-following capabilities. We observe that the model tuned without CoT data fails to generate reasoning chains, even when explicitly prompted to do so. This phenomenon highlights the critical role of \textsc{HiTSR} as a thinking-aware TSR training resource. By integrating high-quality CoT trajectories, our benchmark not only incentivizes robust cross-task reasoning but also synergistically preserves the foundational generative alignment of the backbone model.

\begin{table}[tbp]
\centering
\setlength{\tabcolsep}{2pt}
\renewcommand{\arraystretch}{1.2}
\resizebox{\linewidth}{!}{
\begin{tabular}{c|c|c|c|c}
\toprule
\textbf{Type} & \textbf{Ablation} & \makecell{\textbf{L2}} & \makecell{\textbf{L3}} & \makecell{\textbf{ECG-Grounding}} \\
\midrule
\multirow{3}{*}{ID}  & w/o CoT         & 97.0 & 80.0 & 57.4 \\
                     & joint training    & 95.0 & 84.0 & 61.5 \\
                     \cmidrule(lr){2-5}
                     & \textsc{LLaTiSA} & \textbf{97.0} & \textbf{87.0} & \textbf{62.8} \\
\midrule
\multirow{3}{*}{OOD} & w/o CoT         & 93.3 & 55.0 & 54.3 \\
                     & joint training    & 95.8 & 57.0 & 59.7 \\
                     \cmidrule(lr){2-5}
                     & \textsc{LLaTiSA} & 97.5 & 67.0 & \textbf{62.2} \\
\bottomrule
\end{tabular}
}
\caption{Ablation results. ``w/o CoT'' denotes SFT on the same dataset using only the correct options as supervision during training on L2-L3, ``joint training'' performs a single-stage training regime.}
\label{tab:ablation}
\end{table}

\paragraph{Effectiveness of curriculum learning.} We validate the necessity of a three-stage curriculum for SFT, as opposed to a single-stage regime (i.e., ``joint training'') in which all tasks are pooled and shuffled. As illustrated in \Cref{tab:ablation}, single-stage joint training consistently degrades performance, yielding drops of 2.06\% and 3.45\% in ID tasks for L2 and L3, respectively, and reductions of 1.74\% and 14.93\% under OOD evaluation for L2 and L3. While the joint-training model maintains relatively stable performance on ID scenarios and simpler OOD tasks (L2), it exhibits a significant decline on OOD L3 tasks, suggesting that without curriculum learning, the model struggles to internalize the logic needed for complex semantic reasoning, thereby limiting generalization.

\section{Conclusion}
In this paper, we propose a difficulty-stratified view of TSR and formalize a four-level taxonomy (L1-L4) that characterizes progressive TSR capabilities. We introduce \textsc{HiTSR}, a large-scale hierarchical dataset with over 83k high-fidelity TSR questions spanning L1-L3, constructed with unambiguous labels and verifiable reasoning chains to support both reliable training and evaluation. Building on \textsc{HiTSR}, we present \textsc{LLaTiSA}, a VLM-based TSR model that pairs time series plots with an auxiliary index-value table to bridge qualitative perception and precise numerical evidence, and is trained via a three-stage curriculum aligned with L1-L3. Extensive experiments on diverse OOD benchmarks and real-world scenarios show that \textsc{LLaTiSA} consistently outperforms strong baselines and different modeling strategies, and transfers effectively to domain-specific semantic reasoning, suggesting a practical path toward more robust unified TSRMs.

\section*{Limitations}

Our study primarily focuses on systematic dataset construction and supervised curriculum fine-tuning, leaving the exploration of Reinforcement Learning Fine-Tuning (RFT) on \textsc{HiTSR} as a future direction. Effectively implementing RFT in this domain remains challenging due to the complexity of reward design, which must simultaneously supervise low-level numerical precision and high-level semantic logic across our L1-L4 taxonomy. Future work includes: (i) integrating RL into the hierarchical curriculum to further refine the model's policy; (ii) investigating robust initialization strategies to mitigate cold-start issues.

\section*{Ethics Statement}
All datasets and language models used in this work are publicly available. The \textsc{HiTSR} dataset was constructed from well-established, publicly accessible time series benchmarks and synthetic data generation pipelines, and \textsc{LLaTiSA} is built on the open-sourced model.

\bibliography{acl_latex}

\appendix
\section*{Appendix}

\section{Findings}\label{sec:findings}
\subsection{Model Analysis}\label{subsec:model_findings}
Our research is primarily motivated by a critical appraisal of existing TSR benchmarks. Preliminary evaluations of GPT-4o on local pattern differentiation tasks reveal significant deficiencies in handling foundational TSR operations, echoing the observations in \cite{bedtime}. As illustrated in \Cref{tab:model_bad_case1}, GPT-4o fails to perform accurate numerical grounding when interpreting a visual plot; it struggles to discern the precise relationship between start and end points or characterize the specific rate of change (e.g., confusing abrupt drops with "smooth" decreases). This diagnostic case underscores a pivotal prerequisite: establishing robust numerical read-out capabilities for TSRMs is essential before tackling the higher-order reasoning required in L3 and L4 tasks. Furthermore, our empirical results across both ID (\Cref{tab:baseline}) and OOD (\Cref{tab:zeroshot}) scenarios demonstrate that current LLMs, VLMs, and even specialized TSRMs fall short in foundational L1-L2 tasks, which suggests a fundamental perception-reasoning gap.

\subsection{Dataset Analysis}\label{sec:task_sum}
We systematically categorize existing TSR datasets according to our proposed four-level taxonomy in \Cref{tab:data_sum}, through which we identify several critical deficiencies in current benchmarks. For instance, certain questions exhibit label ambiguity, containing multiple valid options (\Cref{tab:data_bad_case1,tab:data_bad_case2}), while others lack sufficient contextual information for logical reasoning (\Cref{tab:data_bad_case3}). Furthermore, most TSR datasets are designed exclusively for evaluation rather than providing comprehensive training-test splits. Among the limited training data available, the absence of Chain-of-Thought (CoT) annotations is a major drawback, while solely relying on correct labels for training tends to degrade the generative reasoning capabilities of TSRMs.

\section{Data Source}\label{sec:data_source}

\subsection{Synthetic Time Series}
We use the complete time series synthesis pipeline introduced by \cite{chatts}. To avoid ambiguity in options and control question quality, we merge some of the preset attributions to ensure that the remaining attributes do not introduce multiple correct options when used to create local pattern recognition problems. The new attribution pool is summarized in \Cref{tab:syn_ts}.

\subsection{Real-world Time Series}
We utilize real-world time series data for: (i) creating OOD test sets for the L1 task; (ii) constructing semantic understanding subsets.
For the L1 evaluations reported in \Cref{tab:zeroshot}, we select datasets as follows:
\paragraph{ETT (4 subsets)} \cite{informer} contains a group of four subsets of oil temperature and power load collected by electricity transformers from July 2016 to July 2018, with minutes or hourly recorded frequency.

\paragraph{Weather} \cite{weatherd} records meteorological time series with 21 weather indicators collected every 10 minutes
from the Weather Station of the Max Planck Biogeochemistry Institute in 2020.

\paragraph{Exchange Rate} \cite{exchange} includes the daily exchange rates of eight different countries, ranging from 1990 to 2016.

\paragraph{Traffic} \cite{autoformer} encompasses the hourly measures of road occupancy rates obtained from 862 sensors situated in the San Francisco Bay area freeways between January 2015 and December 2016.

\paragraph{Electricity} \cite{autoformer} mainly consists of hourly electricity consumption data from 321 customers.

To facilitate the semantic understanding task, we augment our dataset with large-scale time series repositories, including UTSD \cite{timer}, Monash \cite{monash}, and Time-MMD \cite{timemmd}. These datasets provide rich meta-annotations that serve as the basis for generating ground-truth options. Each time series is subjected to random cropping to generate sub-series, thereby expanding the data volume and enhancing descriptive diversity. Comprehensive statistics of these datasets are provided in \Cref{tab:datasum}.

For each sample, we prompt GPT-5 with the raw time series (represented as textual arrays) and its associated metadata to synthesize scenario-aligned MCQs. The ground truth is derived from the actual data source, while distractors are engineered to be erroneous in terms of numerical precision, temporal pattern perception, or contextual consistency, which correspond to levels L1–L3 in our TSR taxonomy. We further define two semantic evaluation settings: (i) single-scenario, where all options describe the series within the same metadata context, and (ii) multi-scenario, where the LLM is instructed to generate cross-contextual wrong options that deviate from the ground truth.


\begin{table}[t]
\centering
\resizebox{\linewidth}{!}{
\begin{tabular}{l|lc}
\hline
\textbf{Type} & \textbf{attribution} & \textbf{Caption} \\
\hline
\multirow{6}{*}{Global}
& Trend & \makecell{increase, decrease, \\ steady, multi-stage;} \\
\cmidrule(lr){2-3}
& Seasonality & \makecell{sin, square, \\ triangle, none;} \\
\cmidrule(lr){2-3}
& Frequency & high, low; \\
\cmidrule(lr){2-3}
& Noise & noisy, clean; \\
\midrule
\multirow{3}{*}{Local}
& Spike & upward, downward; \\
\cmidrule(lr){2-3}
& Turning point & \makecell{rises then falls, \\ falls then rises;} \\
\midrule
\multirow{4}{*}{Numerical}
& Min/Max & [index, value] pairs; \\
\cmidrule(lr){2-3}
& Start/End & values; \\
\cmidrule(lr){2-3}
& Key event positions & spike, turning point; \\
\bottomrule
\end{tabular}
}
\caption{Overview of synthetic time series attributions.}
\label{tab:syn_ts}
\end{table}

\section{Details of \textsc{HiTSR}}\label{sec:task_intro}
\subsection{Task definition}\label{sec:tasksum}
The full task definition is summarized in \Cref{tab:tasksum}. 

\subsection{Data Verification}
\paragraph{L1: Numerical Read-out.} 
The question-answer (QA) pairs utilized in this stage are completely constructed by strict rule-based annotations based on each time series' attribution pool.
\paragraph{L2: Pattern Perception.} 
For numerical perception, it is similar to L1, which generated the complete QA pairs through an explicit rule-based pipeline. The distractors are created by slightly perturbing the numerical evidence.

For local pattern differentiation, distractors are synthesized using rule-based perturbations, such as numerical manipulation or random attribute sampling from heterogeneous series. To ensure label uniqueness, we perform cross-verification by benchmarking each candidate option against the ground-truth attributes of the series. Subsequently, we use GPT-5 with both the metadata and visual plots of the given time series to annotate the correct answers with CoT reasoning. For quality assurance, the training set only includes samples correctly resolved by GPT-5, while the test set undergoes rigorous human verification to ensure high fidelity.

Regarding global pattern differentiation, we first prompt GPT-5 to generate comprehensive descriptions of the series. Distractors are then constructed by incorporating annotations from disparate sequences. Similarly, the original dataset is processed by GPT-5 to produce reasoning-enhanced ground truth, ensuring the depth and accuracy of the labels.

\paragraph{L3: Semantic Understanding.}
The raw QA pairs are initially synthesized by GPT-5 in a single pass. To ensure data integrity, we implement a cross-validation stage where GPT-5 evaluates the generated pairs against a set of rigorous criteria. This evaluation focuses on two critical dimensions: (i) verifying the ground truth, ensuring that numerical evidence, pattern characterization, and semantic logic are technically sound; and (ii) auditing distractors, where each incorrect option is re-examined to eliminate potential overlaps or counterexamples that might inadvertently align with the time series. Finally, a manual dual-check is performed on the validated candidates to finalize a high-fidelity, real-world TSR dataset.
 
\subsection{Prompt Templates}
\begin{figure*}[t]
  \centering
  \begin{tcolorbox}[promptbox]

  {
  \centering
  \includegraphics[width=0.75\textwidth]{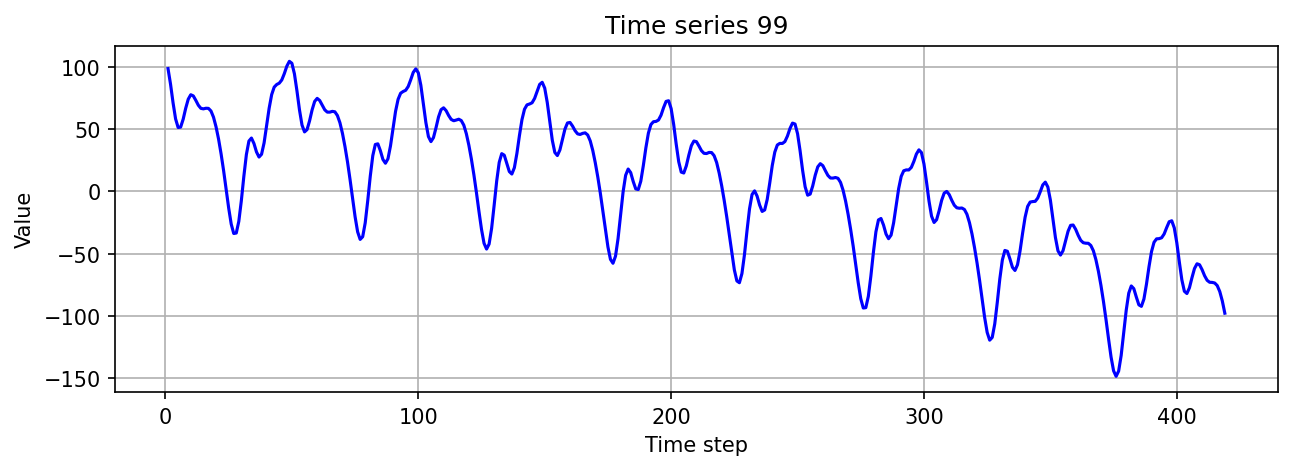}\par
  }
  \vspace{3mm}
  
  \ttfamily\small
You are a time series analysis expert. Generate a clear and brief natural language description of the given time series plot. You can describe its shape, main phases, and key events based on the metadata below. As key numerical characteristics must be included, the precise values represented in the metadata are recommended to be referenced from the coordinate axis division in the plot, and higher precision should be omitted. Only describing approximate values in natural language and do not use `$\sim$' notation. Do not summarize every characteristic in the metadata; generate a concise and specific description in one sentence, like the following example:

This time series shows gradual growth from index 0 to 50, followed by a steady decline until index 100, then exhibits exponential growth peaking at index 280 with a maximum value of 1.259 before sharply dropping.

Metadata:

- Length: 419 points

- Overall trend: decreasing

- Seasonality: seasonality pattern of type `sin' with period of 49.9 points

- maximum value 104.2374 at position 49

- minimum value -148.294 at position 376

- starts at 98.42943844179285 and ends at -97.69269541940011 (overall change 'down')

- Turning points: no significant turning points

- Local events: no major local events

- Trend segments: decrease from index 0 to 418 (values 98.42943844179285→-97.69269541940011)

- volatility level is `high'.

- Cycle stats: peaks between -90.98002956701706 and 104.23737349499228; 

    - valleys between -37.91800010327716 and 102.68065545342499; 
    
    - mean amplitude:-36.82626304378404; 8.4 cycles in total.
  \end{tcolorbox}

  \caption{Prompt template for global pattern annotation.}
  \label{fig:l2_annotate}
\end{figure*}

\begin{figure*}[t]
  \centering
  \begin{tcolorbox}[promptbox]

  {
  \centering
  \includegraphics[width=0.75\textwidth]{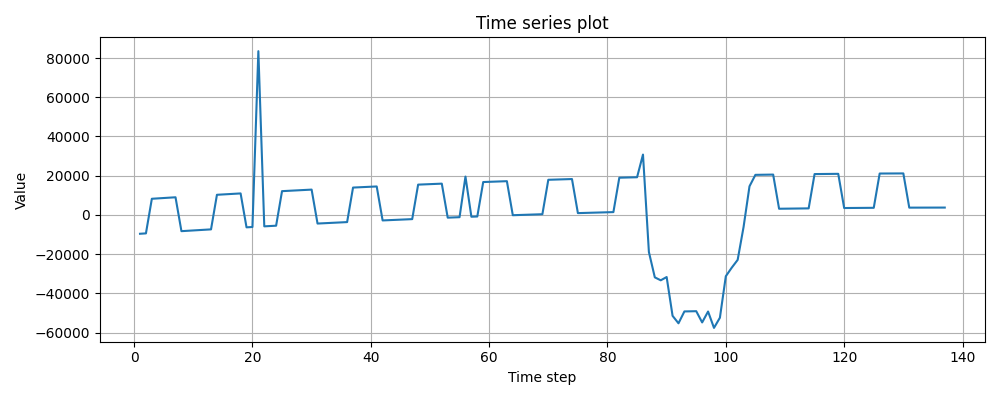}\par
  }
  \vspace{3mm}
  
  \ttfamily\small
You are a time series analysis expert. Carefully analyze the provided time series plot and choose the single best option that most accurately describes the pattern depicted.

Please analyze each option step-by-step, and explain every option whether it matches the given time series plots, and point out any inaccuracies. After you have analyzed all options, output the best answer in the format: ``The correct answer is X''.

Options:

A: The time series exhibits a sinusoidal periodic fluctuation, with an approximate minimum period length of 11.3.

B: The overall trend of the time series is steady, which means no significant increasing or decreasing overall trend.

C: The value at the end of the series is lower than at the start.

D: The time series contains 3 overall trend turning points. Turning points refer to changes in the overall trend direction.
  \end{tcolorbox}

  \caption{Prompt template for generating CoT for L2 tasks.}
  \label{fig:l2_cot}
\end{figure*}

\begin{figure*}[t]
  \centering
  \begin{tcolorbox}[promptbox]

  {
  \centering
  \includegraphics[width=0.75\textwidth]{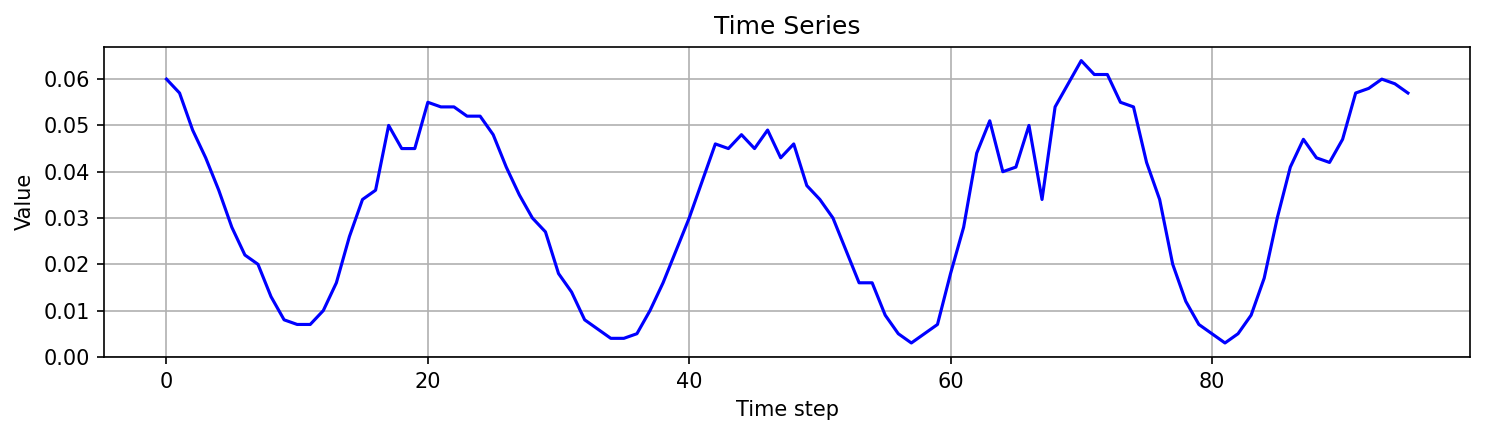}\par
  }
  \vspace{3mm}
  
  \ttfamily\small
You are an expert in Time Series Analysis. Analyze the given numerical time series and generate one high-quality multiple-choice question (MCQ) with a detailed Chain-of-Thought evaluation for each option.

Output \textbf{MUST} be a single, valid JSON object with exactly the following structure:
{
  question: string,
  
  options: {
  
    A: string,
    
    B: string,
    
    C: string,
    
    D: string
  }
  
  chain of thought: {
  
    A: {evaluation: string, error type: none or Pattern/Value/Semantic},
    
    B: {evaluation: string, error type: none or Pattern/Value/Semantic},
    
    C: {evaluation: string, error type: none or Pattern/Value/Semantic},
    
    D: {evaluation: string, error type: none or Pattern/Value/Semantic}
  }
  
  final verification: {
  
    checked values: [list of [index, value] verified],
    
    neighbor consistency: boolean
  },
  
  Final answer: A/B/C/D
}

RULES:
- DO NOT include any text before or after the JSON.

- ALL strings must be properly escaped.

- Use zero-based indexing to verify all [index, value] claims.

- One and only one correct answer; assign it randomly, but ensure others have one clear error:

   - Pattern Error: wrong overall behavior (e.g., says periodic when irregular)
   
   - Value Error: incorrect index-value pair or broken neighbor logic
   
   - Semantic Error: impossible real-world context (e.g., room temp at 8ms)
   
- In options: combine Pattern (P), Values (V), Semantics (S) into one fluent sentence per choice.

- Academic English only. No markdown, no line breaks.

<input-data>

Time series (length=96, sampling interval=hourly,element format: $[$index,value$]$):

$[[0,0.06],[1,0.057],[2,0.049],[3,0.043],[4,0.036],...]$

Source Info: {

frequency: hourly, 

true scenario: This is an hourly measured time series that represents road occupancy rate recorded by San Francisco freeway sensors from 2015 to 2016.

}

</input-data>

Now output ONLY the JSON.
  \end{tcolorbox}

  \caption{Prompt template for generating L3 QA pairs.}
  \label{fig:l3_generate}
\end{figure*}

\begin{figure*}[t]
  \centering
  \begin{tcolorbox}[promptbox]

  {
  \centering
  \includegraphics[width=0.75\textwidth]{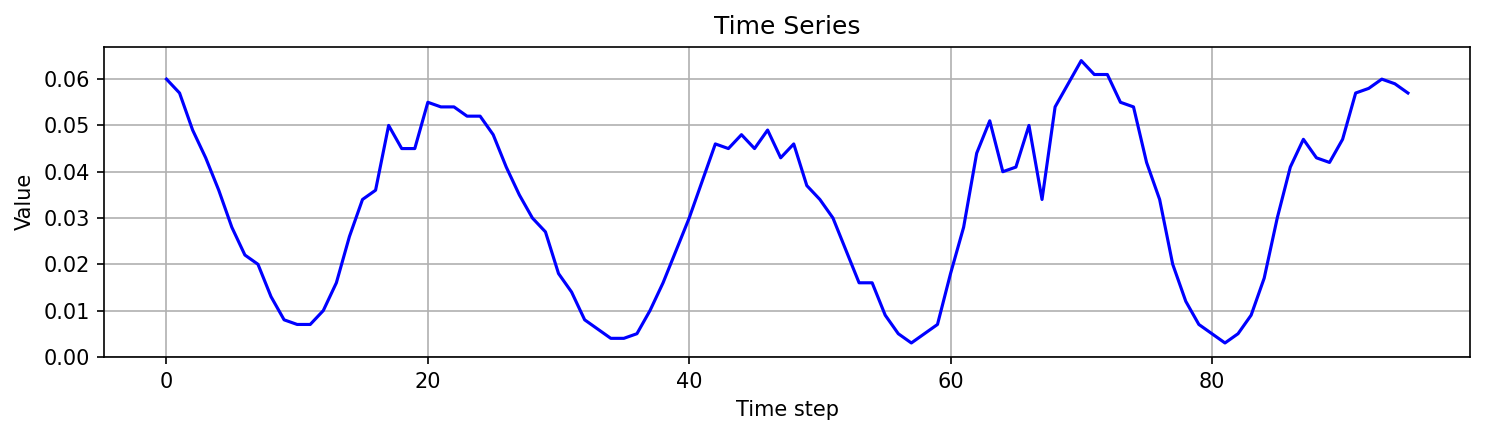}\par
  }
  \vspace{3mm}
  
  \ttfamily\tiny
You are a meticulous Question Examiner. Your task is to evaluate the correctness of the analysis in the ``chain of thought'' section for a multiple-choice time series question. You must verify whether each option's evaluation is strictly accurate — based \textbf{only on the raw time series data}, not on assumptions about real-world timing or formatting conventions.

You can refer to the time series plot to support your evaluation.

The input format is fixed:

timeseries: $[[$index,value$]$,...$]$, 

question: $\{...\}$, 

options: $\{...\}$, 

chain of thought: $\{...\}$, 

final verification: $\{...\}$, 

final answer: X, 

sampling interval: $\{...\}$

Follow these rules RIGOROUSLY. Do NOT use external reasoning that violates these rules.

1.  NO ABSOLUTE TIME INFERENCE ALLOWED — BUT RELATIVE TIME FRAME MUST BE RESPECTED WHEN INTERVAL IS GIVEN

    - NEVER assume that index=0 corresponds to 00:00, 06:00, or any clock time.
    
    - HOWEVER, IF the question specifies a measurement interval (e.g., every 30 minutes), then:
    
        - Elapsed time from start = index $\times$ interval duration in hours.
        
        - Terms like ``morning'', ``evening'', ``peak'', ``night'' must be evaluated for relative plausibility within realistic human activity patterns.
        
        - Example: ``morning rise'' should occur between approximately 6–11 hours from a plausible start; ``evening peak'' between 17–22 hours.
        
        - If two events are labeled as same-day morning and evening, their elapsed times must differ by roughly 6–12 hours, not 30+ hours.
        
        - Most critically: claims of ``daily pattern'' or ``bimodal daily cycle'' require that peaks recur approximately every 48 indices (for 30-min data), or strong evidence of repetition within the series.
        
        - Do NOT accept narrative consistency if temporal spacing makes it physically implausible.

    - You must calculate elapsed times whenever an interval is given, and reject interpretations where:
    
        - A claimed ``same-day''  morning/evening pair spans more than 24 hours,
        
        - Or a ``daily'' pattern lacks repeating structure at ~48-step intervals.

1.1: DAILY PERIODICITY REQUIRES EVIDENCE OF REPETITION

    - For claims involving ``daily'', ``bimodal daily'', ``repeats each day'':
    
        - There must be observable recurrence at intervals close to 48 steps (for 30-min data) or an appropriate multiple.
        
        - At least one full cycle should be visible.
        
        - Values at corresponding phases (e.g., index i and i+48) should be similar within a reasonable tolerance ($\sim \pm 5\%$).
        
        - Absence of such repetition invalidates the claim, regardless of individual peak shapes.

2.  NUMERICAL VERIFICATION WITH CONTEXTUAL TOLERANCE

    - For any claimed value (e.g., ``value at index 2 is 5014.8''):
    
        - Accept it if the actual value rounds to the stated value within one decimal place.
        
        - Specifically: if the option states a number with fewer decimals (e.g., 5014.8), and the true value is 5014.835 $\rightarrow$ this is acceptable.
        
        - Do NOT require exact string match or full precision.
        
        - Only demand exactness if the option uses words like ``exactly'', ``precisely'', or ``equals down to three decimals''.
        
    - This rule applies to all numerical claims, including maxima, minima, and comparisons.

3.  FULL VALUE \& EXTREMUM CHECK

    - For claims involving ``highest'', ``lowest'', ``maximum'', ``minimum'':
    
        - YOU MUST scan all points to confirm.
        
        - But: still apply Rule \#2 — allow rounded reporting.
        
    - Example: option says ``maximum is about 7983.9'' $\rightarrow$ actual max 7983.885 $\rightarrow$ Valid

4.  PATTERN \& TREND: ACCEPT PLAUSIBLE NARRATIVES

    - Do NOT reject an option because another interpretation exists.
    
    - Accept a pattern description if:
    
        - It matches the data trend in the relevant window,
        
        - The terminology is not factually wrong,
        
        - And the shape is realistic under the described scenario.
        
    - Reject only if:
    
        - The trend is opposite (e.g., claims increase but value drops),
        
        - Or generalizes a short fluctuation to the whole series,
        
        - Or uses clearly false periodicity (e.g., ``every 6 hours'' with no recurrence).

5.  SEMANTIC PLAUSIBILITY (REAL-WORLD SENSE)

    - Evaluate whether explanations make sense:
    
        - e.g., ``due to sunrise'' $\rightarrow$ should align with the rising trend at that point
        
        - ``industrial growth'' $\rightarrow$ should show sustained increase, not fluctuations

    - But: never override numerical or pattern correctness due to personal belief.

6.  RESIST CONFIRMATION BIAS

    - Do NOT trust the ``chain of thought'' just because it looks detailed.
    
    - Re-check every claim independently.
    
    - But: if it correctly identifies errors and supports the right answer, mark as valid.

7.  FINAL OUTPUT FORMAT (STRICTLY ENFORCED)

    Output exactly one of:

    ``valid'': ``true'', ``reason'': All analyses are correct.

    valid: false, reason : [Error type]: [Option X] has an issue because [specific reason with index/value or pattern mismatch].

    - Do NOT say ``index 50 is 25 hours from start''.
    
    - Do NOT reject for 5014.835 vs 5014.8 unless the claim is exact.

<inßput-data>
    
</input-data>
  \end{tcolorbox}

  \caption{Prompt template for L3 QA pairs verification.}
  \label{fig:l3_check}
\end{figure*}

The prompt templates used during the construction of \textsc{HiTSR} are summarized as follows:
\begin{itemize}[left=0pt]
\setlength{\itemsep}{0.5pt}
    \item \textbf{L2: Annotation:} \Cref{fig:l2_annotate}.
    \item \textbf{L2: CoT Generation:} \Cref{fig:l2_cot}.
    \item \textbf{L3: QA Generation:} \Cref{fig:l3_generate}.
    \item \textbf{L3: Question Examination:} \Cref{fig:l3_check}.
\end{itemize}

\subsection{Illustration of \textsc{HiTSR}} 
We provide a QA sample for each subtask within \textsc{HiTSR}, which is summarized as follows:
\begin{itemize}[left=0pt]
\setlength{\itemsep}{0.5pt}
    \item \emph{Numerical Read-out:} \Cref{tab:l1_example_1,tab:l1_example_2,tab:l1_example_3,tab:l1_example_4}.
    \item \emph{Pattern Perception:} \Cref{tab:l2_example_1,tab:l2_example_2,tab:l2_example_3}.
    \item \emph{Semantic Reasoning:} \Cref{tab:l3_example}.
\end{itemize}

\subsection{Comparison to Existed TSR Datesets}
The differences among \textsc{HiTSR} and existing TSR datasets are summarized as follows:
\begin{itemize}[left=0pt]
\setlength{\itemsep}{0.5pt}
    \item \textbf{The Enrichment of Verified Chain-of-Thought.} Different from evaluation-only benchmarks (e.g., BEDTime \cite{bedtime}, TimeMCQ2 \cite{mcq2}, and MMTS-Bench \cite{mmtsbench}), \textsc{HiTSR} provides 83k samples enriched with verified Chain-of-Thought (CoT) trajectories. By transforming raw data into high-quality instruction-tuning pairs, we enable TSRMs to learn the "process of reasoning" rather than solely mapping correct answers. Our results (Section 4.5) show that this approach significantly enhances the \textsc{LLaTiSA}'s performance in complex reasoning tasks while preserving the model's general instruction-following capabilities.
    \item \textbf{Unambiguous and Logic-Grounded CoT.} In contrast to the datasets with noisy (see Table 13-14 in the manuscript) or purely semantic annotations, our CoT follows a strict ``Perception-to-Reasoning'' logic (Numerical -> Pattern -> Semantic). Each step is cross-verified against ground-truth data, ensuring that the reasoning path is grounded in numerical reality rather than textual hallucination.
    \item \textbf{Emphasis on Foundational TSR Capabilities.} Most existing studies (e.g., Time-MMD \cite{timemmd}, TSR-Suite \cite{timeomni}) focus on L2-L4 high-level TSR tasks but overlook the verification of a model's numerical read-out ability, especially in long-horizon settings. \textsc{HiTSR} uniquely introduces L1 tasks (precise grounding and localization) as a diagnostic anchor. This ensures that complex TSR abilities are evaluated only after verifying the model's foundational perceptual accuracy.
    \item \textbf{A clear difficulty-Stratified taxonomy for TSR.} Guided by our 4-level taxonomy (grounded in Bloom's Taxonomy \cite{bloom} and Bertin's Levels of Reading \cite{Bertin}), \textsc{HiTSR} acts as a systematic diagnostic framework. Unlike fragmented benchmarks, \textsc{HiTSR}'s stratified design allows for curriculum learning (L1 $\rightarrow$ L3) and enables researchers to pinpoint exactly where a TSRM’s reasoning chain breaks down.
\end{itemize}

\section{Evaluation on the \textsc{HiTSR}}\label{sec:eval_hits}
We evaluate representative baseline models on the \textsc{HiTSR} test set, comprising 100 randomly sampled questions, with results summarized in \Cref{tab:baseline}. Key observations are as follows: First, incorporating temporal indices into textualized numerical arrays significantly enhances the reasoning performance of LLMs; for instance, GPT-4o’s L1 accuracy improves from 6.0\% to 23.0\% with indices, particularly aiding numerical grounding tasks. Second, ChatTS demonstrates a strong capability to estimate approximate values while being poor at mapping the correct index for the target value, achieving a competitive 70.0\% in semantic understanding (L3), which is merely a 1.0\% performance gap compared to the best-performing multimodal GPT-4o.

A consistent performance disparity exists between local and global pattern differentiation, revealing the limitations of single-modality reasoning. While pure vision or text-only models excel in global perception (e.g., GPT-4o reaching 93.0\% and 94.0\% in L2-Global, respectively), they both struggle with local granularity, as evidenced by their stagnant performance in L2-Local (hovering around 55.0\%). This bottleneck is exemplified in \Cref{tab:model_bad_case1}, where pure vision fails to discern precise temporal transitions, such as confusing a sharp step-drop with a ``smooth'' decrease.

Crucially, our results demonstrate a significant performance leap when Vision and Text modalities are integrated, with GPT-4o’s L2-Local accuracy rising to 69.0\%. This synergistic gain highlights the effectiveness of a dual-stream input paradigm—utilizing visual plots for overall pattern recognition and textual sequences for numerical precision. These findings suggest that future TSRMs should prioritize the fusion of visual global context and numerical fine-grained features to overcome the inherent perceptual gaps of isolated modalities.

For the ``Numerical Perception task'' in L2, we introduce subtle numerical perturbations into local pattern differentiation to synthesize challenging distractors. Experimental results in \Cref{tab:baseline} show that Qwen3-VL-8B achieves the state-of-the-art (SOTA) performance of 78.0\% when equipped with both vision and indexed textual series inputs. Besides, we observe a significant performance degradation for Qwen3-8B in the "Text (w/o index)" setting (36.0\%); this is primarily due to its inefficient regressive value retrieval patterns, which often exceed the maximum output length, thus fail to answer the question. Furthermore, the Vision-only modality consistently yields the lower scores across models (e.g., 58.0\% for GPT-4o and 47.0\% for Qwen3-VL). Such a failure of visual heuristics underscores the success of our dataset construction: the ``Numerical Perception'' task necessitates a reliance on precise numerical evidence rather than coarse visual patterns, thereby establishing a more rigorous benchmark for fine-grained TSR.

\section{Training Details}\label{sec:settings}
We provide the training parameter settings and training data proportions of the three-stage SFT in the \Cref{tab:parameter}.

\section{Baseline Methods}\label{sec:baseline}
    
\subsection{Time Series Reasoning Models}
\paragraph{Time-MQA}\cite{timemqa}.
Time-MQA is an LLM-based TSRM that employs LoRA to fine-tune different backbone models on a large-scale TSQA dataset that contains traditional time series analysis and QA tasks.
\paragraph{Time-R1}\cite{timer1}.
Time-R1 is an LLM-based TSRM, which utilizes RFT to transfer the LLM's reasoning abilities to the forecasting task.
\paragraph{ChatTS}\cite{chatts}.
ChatTS is a TS-MLLM built on Qwen2.5-14B-Instruct, which encodes time series via a 5-MLP encoder, training on large-scale synthetic datasets and generalizing to various TSR scenarios.
\paragraph{OpenTSLM}\cite{opentslm}.
OpenTSLM is a domain-specific TS-MLLM built on that adopts two types of architectures. It employs a curriculum training paradigm, which first trains on TS-QA datasets and then transfers to fine-tune on medical-specific TSR tasks.

\subsection{Instruction Templates}
\begin{figure*}[t]
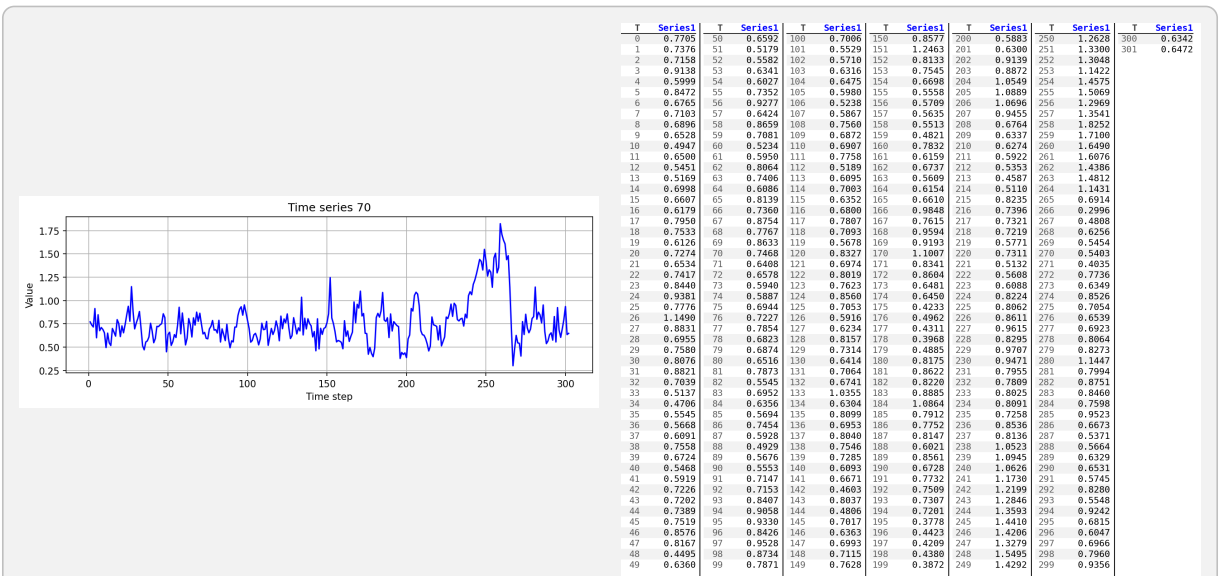

  \centering
  \begin{tcolorbox}[promptbox]
  
  \ttfamily\small
  Given the time series data: 
  
  $[64.234214, 104.221399, 104.478061, 104.747216, 104.972426, 105.241486, 105.503337, ...]$, 
  
  <question>

---
  
  Given the time series data \textbf{(element format[index,value]), start from T = 0}:
  
  $[[0,64.234214], [1,104.221399], [2,104.478061], [3,104.747216], [4,104.972426], [5,105.241486], [6,105.503337], ...]$, 
  
  <question>
  
  \end{tcolorbox}

  \caption{Two types of templates for text models, i.e., ``w/o index'' and ``w/ index''.}
  \label{fig:text_template}
\end{figure*}

\begin{figure*}[t]
  \centering
  \begin{tcolorbox}[promptbox]

  {%
    \centering
    \begin{minipage}{0.49\textwidth}
      \centering
      \includegraphics[width=\linewidth]{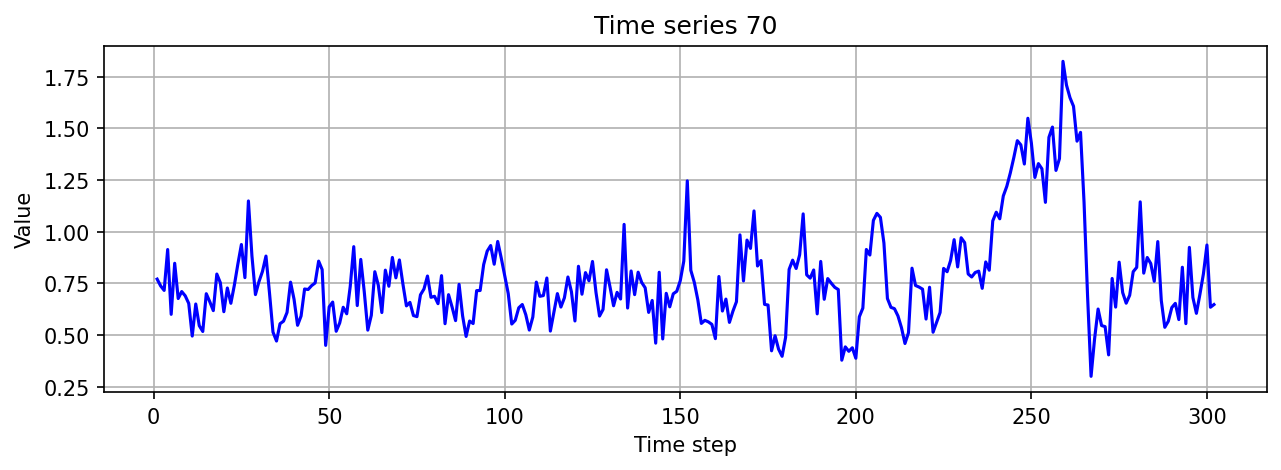}
    \end{minipage}\hfill
    \begin{minipage}{0.49\textwidth}
      \centering
      \includegraphics[width=\linewidth]{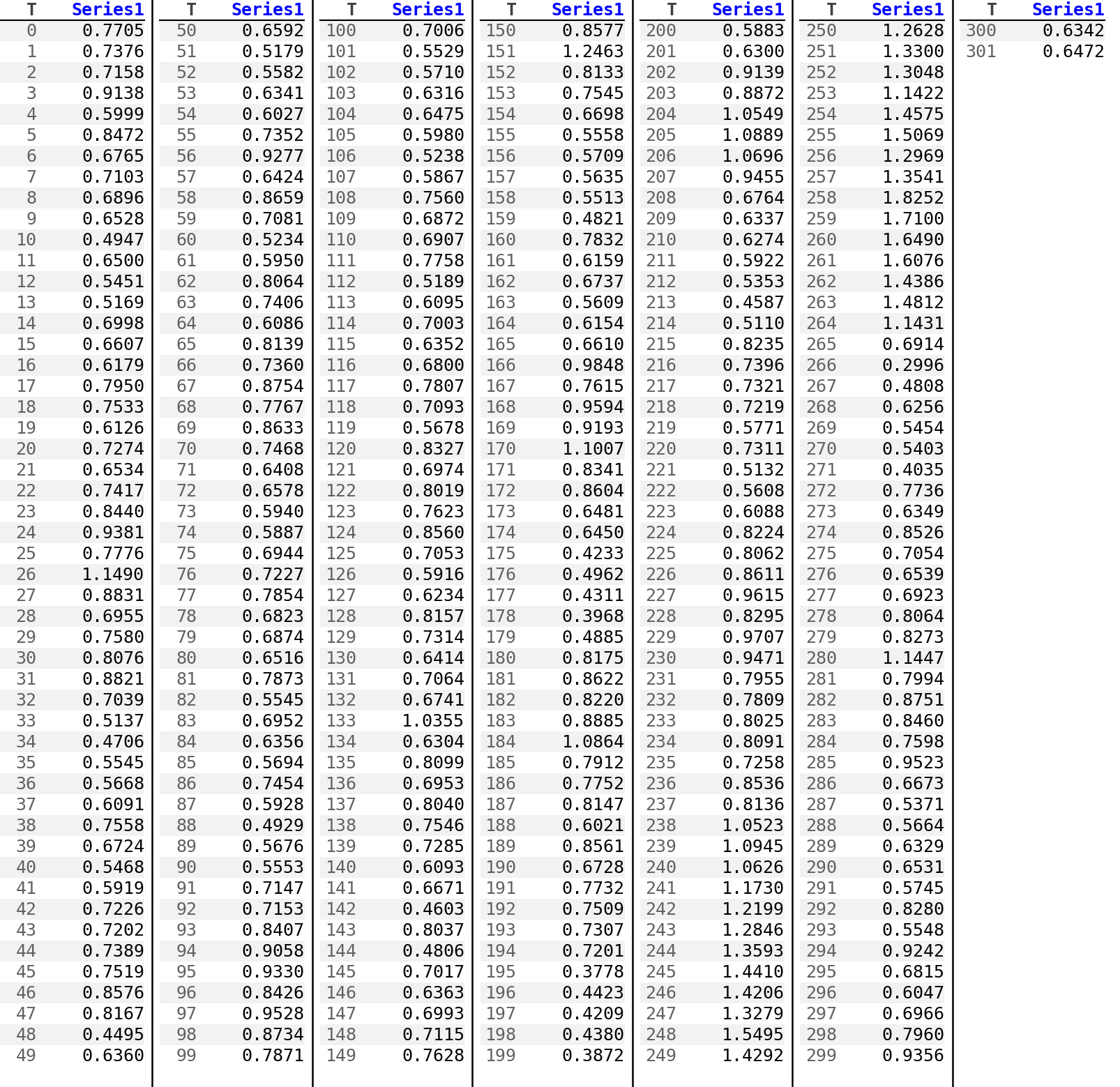}
    \end{minipage}\par
  }
  \vspace{3mm}

  \raggedright
  \ttfamily\small
  <image><image>You must refer to both of the two time series images to answer the question. 
  
  The first image plots the visual trends. The second image is a \textbf{High-Density Numeric Grid} that provides precise values.
  
  \textbf{How to read the second image (Numerical Grid):}
  
  1. \textbf{Layout}: The data is organized in a multi-column newspaper layout. Read the columns from \textbf{left to right}.
  
  2. \textbf{Vertical Flow}: Within each major column, read the data vertically from \textbf{top to bottom}. When a column reaches the bottom, the sequence continues at the top of the next column to the right.
  
  3. \textbf{Structure}: Inside each column block, the \textbf{leftmost sub-column labeled `T'} represents the time index. The subsequent sub-columns (e.g., Series-1) represent the values of different time series.

  <question>

  \end{tcolorbox}

  \caption{Prompt template for numerical table input, e.g., \textsc{LLaTiSA}.}
  \label{fig:2img_template}
\end{figure*}

\begin{figure*}[t]
  \centering
  \begin{tcolorbox}[promptbox]

  {
  \centering
  \includegraphics[width=0.75\textwidth]{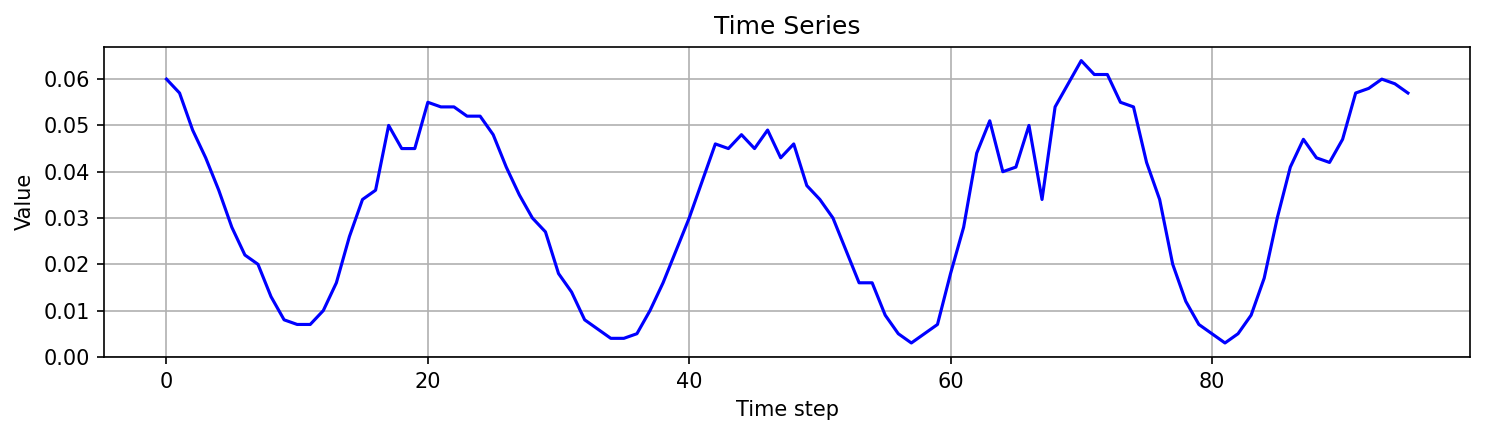}\par
  }
  \vspace{3mm}
  
  \ttfamily\small
  <image>You must refer to the time series' plot and its numerical series to answer the question. Given the time series data: 
  
  $[64.234214, 104.221399, 104.478061, 104.747216, 104.972426, 105.241486, 105.503337, ...]$, 
  
  <question>

---
  
  <image>You must refer to the time series plot and its numerical series to answer the question. Given the time series data \textbf{(element format[index,value]), start from T = 0}:
  
  $[[0,64.234214], [1,104.221399], [2,104.478061], [3,104.747216], [4,104.972426], [5,105.241486], [6,105.503337], ...]$, 
  
  <question>
  
  \end{tcolorbox}

  \caption{Two types of templates for dual-input of image and textual arrays.}
  \label{fig:img_text_template}
\end{figure*}

Our instruction templates for different types of TSRMs are listed as follows:
\begin{itemize}[left=0pt]
\setlength{\itemsep}{0.5pt}
    \item text models:  \Cref{fig:text_template}.
    \item \textsc{LLaTiSA} and VLMs with numerical table image: \Cref{fig:2img_template}.
    \item vision models with textual series input: \Cref{fig:img_text_template}.
    \item For existing TSRMs that are already trained or fine-tuned, we adapt their official prompt templates for the evaluation.
\end{itemize}

\section{Case Study}
We provide valuable cases of how \textsc{LLaTiSA} responds to different TSR tasks in the OOD setting. Cases are summarized as follows:
\begin{itemize}[left=0pt]
\setlength{\itemsep}{0.5pt}
    \item \textbf{L1. Min/Max Grounding.} \Cref{tab:llatisa_l1}.
    \item \textbf{L2. Local Pattern Differentiation.} \Cref{tab:llatisa_l2_local}.
    \item \textbf{L2. Global Pattern Differentiation.} \Cref{tab:llatisa_l2_global}.
    \item \textbf{L3. Semantic Understanding.}
\end{itemize}

\section{AI Assistants Usage}
In this paper, the AI Assistants are only used for text polishing.‘

\section{Theoretical Underpinning of the Difficulty-Stratified Taxonomy}
Our four-level decomposition of TSR is not arbitrary but is grounded in two well-established theoretical frameworks: Bloom's Taxonomy \cite{bloom} (cognitive psychology) and Bertin's Levels of Reading \cite{Bertin} (visual analytic).
\begin{itemize}[left=0pt]
    \item \textbf{Cognitive Reasoning (Bloom's Taxonomy):} Our hierarchy mirrors the progression from Low-Order Thinking Skills (LOTS) to High-Order Thinking Skills (HOTS). Specifically: (i) L1 (Numerical Read-out) maps to Remembering/Retrieving (discrete fact extraction); (ii) L2 (Pattern Perception) maps to Understanding (identifying structural relationships and trends); (iii) L3 (Semantic Reasoning) maps to Applying/Analyzing (contextualizing patterns within domain-specific knowledge); and (iv) L4 (Predictive Inference) maps to Creating (generating future states or logical relationships based on historical temporal data). Recent studies \cite{huber2025llms} have formally advocated for using Bloom's Taxonomy to evaluate the reasoning capabilities of LLMs.
    \item \textbf{Visual Perception (Bertin’s Levels of Reading):} From a data visualization perspective, Jacques Bertin’s classic framework justifies our structural necessity: (i) Elementary Level (L1) focuses on a single data element; (ii) Intermediate/Global Level (L2) focuses on relationships and macroscopic trends across multiple elements; and (iii) Extrapolative Level (L3/L4) involves deducing hidden semantics or future trends beyond the given data.
\end{itemize}
By aligning with these frameworks, \textsc{HiTSR} provides a principled road-map to evaluate whether a TSRM's failure stems from a perceptual bottleneck (L1-L2) or a reasoning deficit (L3-L4). This structured decomposition is necessary for diagnosing the precise cognitive boundaries of TSRMs.

\section{Position of Our Work}

We emphasize that our paper intentionally focuses on TSR-Understanding (L1-L3) rather than TSR-Generation (L4). We argue that these are two fundamentally different paradigms.

\begin{itemize}[left=0pt]
    \item \textbf{Paradigm Distinction.} In the broader multimodal field, understanding models (e.g., Qwen3-VL) and generative model (e.g., the diffusion model) are typically treated as distinct research directions. Integrating L4 (especially numerical forecasting) would necessitate a transition from a Reasoning Model to a Unified Understanding-and-Generation Model, which presents distinct technical challenges (e.g., specialized generative losses) and constitutes a separate research frontier.
    \item \textbf{Diagnostic Purity.} Our primary goal is to provide a diagnostic lens to pinpoint the cognitive boundaries of current TSRMs. By focusing on L1-L3, we ensure a rigorous evaluation of the model's analytical capabilities without the confounding factors of generative precision.
\end{itemize}

\section{Evaluation on L4 Task}
We have conducted evaluations on the L4 (Predictive Inference) task and incorporated L4 into \textsc{HiTSR}. We then trained \textsc{LLaTiSA} on this subset and evaluated its out-of-distribution (OOD) performance on real-world time series data.

\subsection{Task Introduction}
Given a time series segment $x$, the model is tasked with a Multiple Choice (MC) question to identify the unique chronological successor from a pool of four candidate patches $y_1, y_2, y_3, y_4$. While the correct successor follows $x$ immediately in the original series, the negative candidates are sampled from distinct time series with diverse characteristics. This setup evaluates the model's ability to extrapolate underlying temporal dynamics and distinguish the true evolution of a series from statistically distinct alternatives.

\begin{table}[t]
\centering
\resizebox{\linewidth}{!}{
\begin{tabular}{lll}
\toprule
\textbf{Modality} & \textbf{Model} & \textbf{Acc\%} \\
\midrule
\multirow{5}{*}{Text (w/o index)} & GPT-4.1 & 79.1 \\
 & GPT-4o & 75.6 \\
 & Claude-3.5-Sonnet & 82.2 \\
 & Qwen3-8B & 67.1 \\
 & LLaMA3.1-8B & 41.0 \\
\midrule
\multirow{4}{*}{Vision (plot + num)} & GPT-4o & 78.3 \\
 & Qwen3-VL-8B & 42.1 \\
 & \textsc{LLaTiSA} (L1$\rightarrow$L2$\rightarrow$L3) & 54.2 \\
 & \textsc{LLaTiSA} (L1$\rightarrow$L2$\rightarrow$L3$\rightarrow$L4) & \textbf{83.3} \\
\bottomrule
\end{tabular}
}
\caption{Performance comparison on \textbf{out-of-distribution datasets} in L4 (Generative Inference).}
\label{tab:l4_results}
\end{table}

\subsection{Dataset Construction}
We extended our synthetic generation pipeline to ensure that L4 tasks are both challenging and unambiguous. To prevent ``shared ground truth'' (where multiple options could be plausible), we implemented a multi-stage filtering protocol for negative candidates:
\begin{itemize}[left=0pt]
    \item \textbf{Statistical Uniqueness.} We filter out candidates with high Pearson Correlation ($\geq$ 0.8) or small Euclidean Distance to the ground truth to ensure distinctiveness.
    \item \textbf{Strategic Truncation.} To avoid introducing excessive ambiguity, we ensure the split point between $x$ and $y$ does not occur during high-variance events, such as mid-spike or at sharp turning points. Splitting at such critical junctures may cause multiple plausible answers; by selecting structurally stable partition points, we ensure the ground truth is the only logically consistent successor.
    \item \textbf{Trend Smoothness.} We use a weighted continuation score, which incorporates gradient consistency and tail-to-head difference to verify that the positive sample is the most semantically and numerically plausible successor.
    \item \textbf{Mutual Exclusion.} Negative candidates are cross-checked to ensure low mutual correlation, preventing redundant options within a single question.
\end{itemize}

\subsection{Transition to Vision Modalities}
To align with the dual-view strategy of \textsc{LLaTiSA}, we transform the L4 predictive task into a dual-image input format. Specifically, the visualized patterns are rendered as a $2\times2$ grid of subplots, each representing one of the four options. Within each subplot, the lookback context and the candidate prediction patch are visualized in blue and red, respectively, allowing the model to perform visual comparative analysis of temporal continuity. Complementing this, a precision-calibrated numerical table provides exact values for the lookback window. This dual-view setup ensures the model can leverage global trend intuition from the plots while maintaining the fine-grained numerical grounding necessary for rigorous successor identification.

\subsection{Experiment Results}
Following this rigorous protocol, we expanded our curriculum to include a fourth stage (L1$\rightarrow$L2$\rightarrow$L3$\rightarrow$L4). We trained \textsc{LLaTiSA} on 4,993 L4-specific questions for 2 epochs (learning rate $1 \times 10^{-5}$) and evaluated its OOD performance on the real-world TRQA \cite{trqa} subset. As results shown in \cref{tab:l4_results}, we summarize key observations as follows:
\begin{itemize}[left=0pt]
    \item \textbf{Surpassing Text-only TSRMs.} \textsc{LLaTiSA} achieves superior performance compared to both closef-source and open-source TSRMs with textual time series input settings. These results further underscore the effectiveness of \textsc{LLaTiSA}'s dual-view multimodal paradigm, high-quality TSR dataset (\textsc{HiTSR}), and multi-stage curriculum fine-tuning.
    \item \textbf{Comparison with General VLMs.} \textsc{LLaTiSA} significantly outperforms its backbone model Qwen3-VL-8B. This discrepancy highlights the necessity of our dual-view representation and multi-stage training on the difficulty-stratified TSR datasets.
    \item \textbf{Efficacy of the L4 Curriculum Stage.} The transition from the L3-trained model (54.2\%) to the L4-trained model (83.3\%) reveals a substantial 29.1\% improvement. This confirms that while foundational understanding (L1-L3) provides a necessary base, specific predictive inference training (L4) is essential for generalizing the TSR capability from understanding to the generative tasks.
    \item \textbf{Robustness in OOD Scenarios.} Since the evaluation was conducted on real-world data from TRQA (which differs significantly from our synthetic training distributions), the high accuracy (83.3\%) serves as strong evidence of \textsc{LLaTiSA}'s cross-domain generalization. The model has learned the underlying principles of temporal continuity rather than memorizing specific patterns.
\end{itemize}

Overall, these results provide a definitive answer to the reviewers' concerns: \textsc{LLaTiSA} is not only a diagnostic tool for understanding but also has the potential to generalize to L4 predictive tasks.

\begin{table}[t]
\centering
\resizebox{\linewidth}{!}{
\begin{tabular}{ll}
\toprule
\textbf{Ablation} & \textbf{Acc\%} \\
\midrule
only L1 & 16.9 \\
only L2 & 23.0 \\
only L3 & 43.0 \\
\midrule
L1+L2 & 21.4 \\
L1+L2+L3 & 36.8 \\
L1+L2+L3+L4 & 75.4 \\
\midrule
L1$\rightarrow$L2 & 32.8 \\
L1$\rightarrow$L2$\rightarrow$L3 & 54.2 \\
\midrule
\textsc{LLaTiSA} (L1$\rightarrow$L2$\rightarrow$L3$\rightarrow$L4) & \textbf{83.3} \\
\bottomrule
\end{tabular}
}
\caption{Ablation study on the L4 task. All experiments are conducted on OOD datasets.}
\label{tab:l4_ablations}
\end{table}

\subsection{Ablation Study}
To further investigate the synergy between our difficulty-stratified taxonomy and the curriculum training regime, we conducted additional ablation studies on the L4 task. We compared \textsc{LLaTiSA} against single-stage training and joint multi-task learning to verify if foundational levels (L1–L3) are prerequisites for predictive inference (L4). Here, ``+'' refers to joint training on shuffled datasets from multiple levels (e.g., ``L1+L2''), while ``$\rightarrow$'' denotes a sequential curriculum regime (e.g., ``L1$\rightarrow$L2''). Evaluation results (\cref{tab:l4_ablations}) underscore the critical importance of our sequential curriculum: single-stage training fails on L4 tasks without the foundational grounding provided by L1–L3. Notably, \textbf{LLaTiSA} outperforms joint training (``L1+L2+L3+L4'') by 7.9\%, confirming that a step-by-step mastery of temporal primitives and semantics effectively improves the performance of complex predictive inference. This validates that our difficulty-stratified taxonomy acts as a necessary cognitive scaffolding, ensuring robust generalization in complex OOD scenarios.

\section{Expansion of  Experiments}
\subsection{Ablation Study}
To justify the necessity of the multi-stage SFT, we provide intermediate evaluations of \textsc{LLaTiSA} across different training stages on both ID and OOD benchmarks (Table below). Here, ``L1 + L2'' refers to joint training on shuffled datasets from L1-L2, while ``L1 $\rightarrow$ L2'' denotes a two-stage curriculum fine-tuning regime. The evaluation results (\cref{tab:extended_albations}) reveal three critical insights that justify the necessity of our three-stage curriculum:
\begin{itemize}[left=0pt]
    \item As shown in the ``Only L2'' row, although the model masters pattern perception, its L1 performance significantly collapses. This proves that high-level TSR does not naturally "backwards inherit" foundational perception. The L1 stage is essential to ``anchor'' the TSRM's grounding capability.
    \item While training on L2 alone provides some reasoning capability, the performance on L3 improves when L1 and L2 are combined (from 66.0\% to 70.0\% ID). The full curriculum (\textsc{LLaTiSA}) achieves the highest L3 score (87.0\% ID, 67.0\% OOD).
    \item \textsc{LLaTiSA} consistently outperforms ``L1 $\rightarrow$ L2'' and ``L1 + L2'' strategies, indicating the effectiveness of the training on the L3 part and multi-stage curriculum regime, respectively.
\end{itemize}
In conclusion, the curriculum regime is necessary for \textsc{LLaTiSA} because it ensures to maintain a balanced training across different cognitive levels, which a single-stage training on high-level data fails to provide.

\subsection{Baseline Comparison}
We further evaluate ChatTime \cite{chattime} on the datasets we used across \cref{tab:baseline} and explain the differences between TSRMs and Time-LLM \cite{timellm}. We clarify our evaluation results (\cref{tab:extended_exp}) and the architectural distinctions below:

\begin{itemize}[left=0pt]
    \item \textbf{Evaluation on ChatTime.} We evaluated ChatTime on our OOD benchmarks. As shown in \cref{tab:extended_exp}, ChatTime performs poorly across all levels. Notably, it fails to adhere to instruction-defined output formats (e.g., the ``index-value'' requirement for L1 tasks). Furthermore, as ChatTime only accepts a single series, we had to manually concatenate multiple series and specify segment lengths to enable it to finish the ``Series Comparison'' (L3) task.
    \item \textbf{Distinction with Time-LLM.} \textsc{LLaTiSA} significantly outperforms its backbone model Qwen3-VL-8B. This discrepancy highlights the necessity of our dual-view representation and multi-stage training on the difficulty-stratified TSR datasets.
\end{itemize}

\begin{table*}[htbp]
\centering
\setlength{\tabcolsep}{8pt}
\renewcommand{\arraystretch}{1.2}
\resizebox{\textwidth}{!}{
\begin{tabular}{cc|cc|cc|c}
\toprule
\multirow{4}{*}{\textbf{Model}} & \multirow{4}{*}{\textbf{Modality}} & \multicolumn{2}{c|}{\textbf{L1 }\emph{Numerical Read-out}} & \multicolumn{2}{c|}{\textbf{L2 }\emph{Pattern Perception}} & \textbf{L3 }\emph{Semantic Reasoning} \\
\cmidrule(lr){3-7}
& & \multicolumn{2}{c|}{\textbf{Min \& Max Localization}} & \textbf{Local} & \textbf{Global} & \textbf{Series Comparison} \\
\cmidrule{3-7}
& & Acc\% & half-Acc\% & Acc\% & Acc\% & Acc\% \\
\midrule
ChatTime & Time Series & 0.0 & 0.0 & 9.8 & 14.2 & 6.0 \\
\midrule
\textsc{LLaTiSA} & Vision(plot+num) & \textbf{86.8} & \textbf{96.0} & \textbf{75.6} & \textbf{97.5} & \textbf{67.0} \\
\bottomrule
\end{tabular}
}
\caption{Performance comparison on \textbf{out-of-distribution datasets} across L1-L3 with ChatTime.}
\label{tab:extended_exp}
\end{table*}

\begin{table*}[htbp]
\centering
\setlength{\tabcolsep}{8pt}
\renewcommand{\arraystretch}{1.2}
\resizebox{\textwidth}{!}{
\begin{tabular}{cc|cc|cc|c}
\toprule
\multirow{4}{*}{\textbf{Model}} & \multirow{4}{*}{\textbf{Modality}} & \multicolumn{2}{c|}{\textbf{L1 }\emph{Numerical Read-out}} & \multicolumn{2}{c|}{\textbf{L2 }\emph{Pattern Perception}} & \textbf{L3 }\emph{Semantic Reasoning} \\
\cmidrule(lr){3-7}
& & \multicolumn{2}{c|}{\textbf{Min \& Max Localization}} & \textbf{Local} & \textbf{Global} & \textbf{Series Comparison} \\
\cmidrule{3-7}
& & Acc\% & half-Acc\% & Acc\% & Acc\% & Acc\% \\
\midrule
\multirow{4}{*}{InternVL3-8B} 
 & Vision(plot) & - & - & 49.0 & 82.0 & 46.0 \\
 & Vision + Text(w/o index) & 16.8 & 59.2 & 41.8 & 85.8 & 56.0 \\
 & Vision + Text(w/ index) & 36.0 & 72.4 & 44.0 & 87.5 & 55.0 \\
 & Vision (plot+num) & 6.2 & 36.8 & 27.6 & 88.3 & 58.0 \\
\midrule
\multirow{4}{*}{Gemma3-12B} 
 & Vision(plot) & - & - & 64.4 & 91.7 & 51.0 \\
 & Vision + Text(w/o index) & 24.0 & 73.2 & 59.8 & 90.0 & 49.0 \\
 & Vision + Text(w/ index) & 51.4 & 81.0 & 61.2 & 92.5 & 53.0 \\
 & Vision (plot+num) & 7.8 & 43.4 & 55.3 & 95.8 & 41.0 \\
\midrule
\textsc{LLaTiSA} & Vision(plot+num) & \textbf{86.8} & \textbf{96.0} & \textbf{75.6} & \textbf{97.5} & \textbf{67.0} \\
\bottomrule
\end{tabular}
}
\caption{Performance comparison on \textbf{out-of-distribution datasets} across L1-L3 with superior VLMs. The ``-'' denotes that the single plot input cannot handle L1 tasks. }
\label{tab:extended_vlm}
\end{table*}

\begin{table}[t]
\centering
\resizebox{\linewidth}{!}{
\begin{tabular}{lllll}
\toprule
\textbf{Type} & \textbf{Ablation} & \textbf{L1} & \textbf{L2} & \textbf{L3} \\
\midrule
\multirow{6}{*}{ID} 
& only L1 & \textbf{98.2} & 87.5 & 47.0 \\
& only L2 & 14.2 & 96.5 & 66.0 \\
\cmidrule{2-5}
& L1 + L2 & 98.0 & 96.0 & 70.0 \\
& L1 $\rightarrow$ L2 & 97.8 & 96.5 & 72.0 \\
\cmidrule{2-5}
& \textsc{LLaTiSA} (L1 $\rightarrow$ L2 $\rightarrow$ L3) & 98.0 & \textbf{97.0} & \textbf{87.0} \\
\midrule
\multirow{6}{*}{OOD} 
& only L1 & \textbf{97.2} & 87.5 & 47.0 \\
& only L2 & 10.0 & 96.5 & 52.0 \\
\cmidrule{2-5}
& L1 + L2 & 96.0 & 96.0 & 53.0 \\
& L1 $\rightarrow$ L2 & 96.0 & \textbf{97.5} & 56.0 \\
\cmidrule{2-5}
& \textsc{LLaTiSA} (L1 $\rightarrow$ L2 $\rightarrow$ L3) & 96.0 & \textbf{97.5} & \textbf{67.0} \\
\bottomrule
\end{tabular}
}
\caption{Extended ablation study results on the curriculum learning strategy.}
\label{tab:extended_albations}
\end{table}

\subsection{Expansion of VLM Baselines}
To provide a more comprehensive evaluation, we have extended our benchmarks to include InternVL3-8B \cite{internvl3} and Gemma3-12B \cite{gemma3} across various visual input modalities. Following the same experimental settings introduced in Section 4.1, additional results are shown in \cref{tab:extended_vlm}.

Empirical results demonstrate that \textsc{LLaTiSA} consistently and significantly outperforms various VLM-based baselines of a comparable parameter scale, regardless of the input modalities or representation strategies employed to numerical series. Notably, these newly incorporated VLM-based models exhibit a similar performance scale to the Qwen3-VL baselines reported in \cref{tab:baseline}, which all fall significantly short of \textsc{LLaTiSA} in both L1 numerical grounding and L3 semantic reasoning tasks. This further validates that our performance gains stem from both the specialized dual-view architecture and three-stage curriculum training paradigm.

\begin{table}[t]
\centering
\resizebox{\linewidth}{!}{
\begin{tabular}{lrrr}
\toprule
\textbf{Task} & \textbf{GPT-5} & \textbf{Qwen3} & \textbf{Cohen's Kappa} \\
\midrule
L2 (Local) & 16,121 / 20,000 & 16,149 / 20,000 & 0.865 \\
\midrule
L2 (Global) & 24,654 / 25,000 & 24,572 / 25,000 & 0.847 \\
\midrule
L3 (Semantic) & 3,321 / 4,000 & 3,378 / 4,000 & 0.831 \\
\bottomrule
\end{tabular}
}
\caption{Agreement results and Cohen's Kappa for different tasks.}
\label{tab:cohens_kappa}
\end{table}

\section{Data Quality Assurance}
\subsection{Human Verification}
To ensure the data quality, we employed a hybrid verification pipeline combining deterministic scripts and human expert audits:

\begin{itemize}[left=0pt]
    \item \textbf{L1.}\emph{Numerical Read-out}: 3,000/30,000 (10\%). Given the objective nature of read-out tasks, we primarily utilized deterministic scripts for data verification.
    \item \textbf{L2.}\emph{Pattern Perception}: 5,100/50,703 (10\%). This subset includes 1,600 Local, 2,500 Global verified by human experts, and 1,000 Numerical Perception samples verified by deterministic scripts.
    \item \textbf{L3.}\emph{Semantic Reasoning}: 3,121/3,121 (100\%). Due to the high cognitive complexity of L3, every single sample underwent manual human verification to ensure reasoning integrity.
\end{itemize}
Our experts followed a three-fold protocol: (1) Numerical Faithfulness to the source series, (2) Logical Consistency of the CoT trajectories, and (3) Ground-truth Correctness. To evaluate the reliability of our protocol, two independent annotators re-verified a random subset of 500 samples from L2 and L3. \textbf{We achieved a Cohen’s Kappa of 0.853, signifying ``almost perfect agreement.''} This score demonstrates that our verification criteria are objective and consistent across different human experts.

\subsection{Cross-Model Validation}
To ensure the annotations are objective and free from model-specific biases, we conducted a cross-model validation using Qwen3-235B-A22B-Instruct (Qwen3) on the subset of \textsc{HiTSR} which contains verifiable CoT trajectories. The following table reports the successfully validated instance counts (number of passed samples relative to the total evaluated) and the inter-annotator agreement scores (Cohen's Kappa calculated between GPT-5 and Qwen3). \textbf{All subsets across L2-L3 achieve ``almost perfect agreement.'' with an average score of 0.848.} Combined with the full human verification on L3, such cross-model evidence demonstrates that advanced language models converge on a consistent standard of validity, effectively serving as a robust, model-agnostic filter to maximize the efficiency and reliability of subsequent human verification.

\begin{table}[t]
\centering
\resizebox{\linewidth}{!}{
\begin{tabular}{l|ll}
\toprule
\textbf{Training Stage} & \textbf{Hyparameter} & \textbf{Assignment} \\
\midrule
\multirow{5}{*}{Shared}
& Base model & Qwen3-VL-8B-Instruct \\
& Computing environment & 8*H20 GPUs \\
& Optimizer & AdamW \\
& Learning rate scheduler & Cosine \\
& Weight decay & 0.1 \\
\midrule
\multirow{6}{*}{Stage-1}
& Epochs & 1 \\
& Warm-up steps & 50 \\
& Batch size per device & 2 \\
& Gradient accumulation steps & 2 \\
& Learning rate & 1e-5 \\
& \#Samples & 2,7000 \\
\midrule
\multirow{6}{*}{Stage-2}
& Epochs & 2 \\
& Warm-up steps & 50 \\
& Batch size per device & 2 \\
& Gradient accumulation steps & 2 \\
& Learning rate & 1e-5 \\
& \#Samples & 4,5000 \\
\midrule
\multirow{6}{*}{Stage-3}
& Epochs & 2 \\
& Warm-up steps & 50 \\
& Batch size per device & 1 \\
& Gradient accumulation steps & 1 \\
& Learning rate & 1e-5 \\
& \#Samples & 2,700 \\
\midrule
\multirow{6}{*}{Stage-ECG}
& Epochs & 2 \\
& Warm-up steps & 50 \\
& Batch size per device & 1 \\
& Gradient accumulation steps & 1 \\
& Learning rate & 1e-5 \\
& \#Samples & 3,0000 \\
\bottomrule
\end{tabular}
}
\caption{Hyperparameter settings for each training stage.``\#Samples'' denotes the number of training items within the corresponding subsets of the \textsc{HiTSR} dataset.}
\label{tab:parameter}
\end{table}

\begin{table*}[tbp]
\centering
\setlength{\tabcolsep}{6pt}
\renewcommand{\arraystretch}{1.15}
\resizebox{\linewidth}{!}{
\begin{tabular}{l|l|c|c}
\toprule
\textbf{Data} & \textbf{Meta Information} & \textbf{Crop Length} & \textbf{Source} \\
\midrule
ETTh (1/2) & \makecell{Hourly measured time series of the transformer oil temperature \\ in the Electricity Transformer Temperature dataset.} & $[96, 192, 336, 720]$ & \cite{informer} \\
ETTm (1/2) & \makecell{15-minute measured time series of the transformer oil temperature \\ in the Electricity Transformer Temperature dataset.} & $[96, 192, 336, 720]$ & \cite{informer} \\
Weather & \makecell{10-minute measured time series of a weather-related target variable, \\ commonly treated as air temperature, from a weather station dataset.} & $[96, 192, 336, 720]$ & \cite{weatherd} \\
Exchange Rate & \makecell{1-daily measured time series of the Singapore exchange rate.} & $[60, 120, 240, 360]$ & \cite{exchange} \\
Traffic.A & \makecell{Hourly measured time series of the road occupancy rate \\ recorded by San Francisco freeway sensors from 2015 to 2016.} & $[96, 192, 336, 720]$ & \cite{autoformer} \\
Electricity & \makecell{Hourly measured time series of the electricity consumption of \\ one customer selected from 321 customers collected from 2012 to 2014.} & $[96, 192, 336, 720]$ & \cite{autoformer} \\
Illness & \makecell{Weekly measured time series covering from 2002 to 2021 of patient counts \\ reported in influenza-related surveillance statistics.} & $[24, 36, 52, 104]$ & \cite{illness} \\
\midrule
Economy & \makecell{Monthly measured time series that represents the international trade balance.} & $[96, 120, 180, 447]$ & \multirow{7}{*}{\cite{timemmd}}  \\
Energy & \makecell{1-weekly measured time series that represents gasoline prices.} & $[full, 120, 240, 480]$ & \\
Environment & \makecell{1-daily measured time series that represents the air quality index.} & $[full, 240, 480, 576]$ & \\
Health US & \makecell{1-weekly measured time series that represents influenza patients.} & $[full, 120, 240, 480]$ & \\
Security & \makecell{Quantifying pattern attributes (e.g., amplitude, slope).} & $[full, 96, 120, 240]$ & \\
SocialGood & \makecell{A monthly measured time series that represents the unemployment rate.} & $[96, 120, 180, 360, 512]$ & \\
Traffic.B & \makecell{A monthly measured time series that represents travel volume.} & $[96, 120, 180, 360, 480]$ & \\
\midrule
Electricity Demand & \makecell{This is a time series representing electricity demand in Victoria, \\ Australia, for 2014, measured at 30-minute intervals.} & $[96, 192, 336, 720]$ & \multirow{14}{*}{\makecell{\cite{timer} \\ \cite{monash}}} \\
Australia Rainfall & \makecell{This is an hourly measured time series of \\ the temperature change in some part of Australia.} & $[96, 192, 336, 720]$ \\
London Smart Meters & \makecell{This is a half-hourly measured time series that represents of \\ the energy consumption readings of London households in kWh.} & $[96, 192, 336, 720]$ \\
Wind Farms & \makecell{This is a time series representing the wind power production of \\ an Australian wind farm, measured every 4 seconds.} & $[128, 256, 512, 1024]$ \\
Benzene Concentration & \makecell{This is an hourly measured time series of the benzene concentration.} & $[96, 192, 336, 720]$ \\
Atrial Fibrillation & \makecell{This is a time series representing atrial fibrillation, measured every 0.008 seconds.} & $[256, 512, 720]$ \\
IEEEPPG & \makecell{This is a time series of heart rate estimation measured every 0.008 seconds.} & $[128, 256, 512, 720]$ \\
SDWPF & \makecell{This is a time series representing the active power output (Patv, kW) \\ of a wind turbine in a single wind farm in China, measured every 10 minutes.} & $[60, 120, 240, 360]$ \\
Wind Speed & \makecell{This is a time series representing the nacelle-measured wind speed (Wspd, m/s) \\ for a wind turbine in a single wind farm in China, measured every 10 minutes.} & $[60, 120, 240, 360]$ \\
Saugeen Day & \makecell{This is a time series representing the mean daily flow of \\ the Saugeen River at Walkerton in cumecs (m³/sec), measured daily.} & $[60, 180, 365, 730]$ \\
Starlight & \makecell{This is a synthetic time series that simulates the \\ brightness of a celestial object in the night sky, measured daily.} & $[60, 180, 365, 730]$ \\
KDD Cup 2018 & \makecell{This is a time series representing the PM2.5 concentration \\ in Beijing from 2017 to 2018, measured hourly.} & $[72, 120, 240, 360]$ \\
Pedestrian Counts & \makecell{This is a time series representing the pedestrian count in Melbourne city, \\ measured hourly, starting in May 2009.} & $[72, 120, 240, 360]$ \\
Kaggle Web Traffic & \makecell{This is a time series representing the daily views \\ of a random Wikipedia article, measured daily.} & $[60, 120, 180, 365]$ \\
\bottomrule
\end{tabular}
}
\caption{Summary of real-world time series data utilized in \textsc{HiTSR}. ``Crop Length'' denotes the random cropping length of the subseries utilized to create an MCQ. ``$full$'' denotes using the full time series. The ``Meta Information'' is utilized for creating L3 QA pairs.}
\label{tab:datasum}
\end{table*}

\begin{table*}[tbp]
\centering
\setlength{\tabcolsep}{6pt}
\renewcommand{\arraystretch}{1.15}
\resizebox{\linewidth}{!}{
\begin{tabular}{l|l|c|c|c}
\toprule
\textbf{Subtask} & \textbf{Capability Target} & \textbf{Data} & \textbf{\#Samples} & \textbf{Length Range} \\
\midrule
\rowcolor[gray]{0.95} \multicolumn{5}{l}{\textit{L1: Numerical Read-out}} \\
Min/Max Grounding        & Index-aware value grounding and verification            & Synthetic & 10,000 & $[64, 512]$ \\
Multi-series Comparison  & Cross-series grounding under aligned time/value queries  & Synthetic & 10,000 & $[64, 256]$ \\
Start/End Comparison     & Boundary grounding (initial vs.\ terminal states)        & Synthetic & 5,000 & $[64, 512]$ \\
Subseries Localization   & Interval-based retrieval with temporal alignment         & Synthetic & 5,000 & $[64, 512]$ \\
\midrule
\rowcolor[gray]{0.95} \multicolumn{5}{l}{\textit{L2: Pattern Perception}} \\
Local Pattern Differentiation & Multi-scale morphology and trend alignment    & Synthetic & 16,098 & $[64, 256]$ \\
Global Pattern Differentiation & Multi-scale morphology and trend alignment    & Synthetic & 24,605 & $[64, 1024]$ \\
Numerical Perception & Quantifying pattern attributes (e.g., amplitude, slope) & Synthetic & 10,000 & $[64, 256]$ \\
\midrule
\rowcolor[gray]{0.95} \multicolumn{5}{l}{\textit{L3: Semantic Reasoning}} \\
Semantic Understanding & Contextual interpretation and decision-making over series evidence & Real-world & 3,121 & $[24, 2048]$ \\
\bottomrule
\end{tabular}
}
\caption{Task overview of \textsc{HiTSR} (L1-L3). Each level groups multiple subtasks.}
\label{tab:tasksum}
\end{table*}

\begin{table*}[htbp]
\centering
\setlength{\tabcolsep}{8pt}
\renewcommand{\arraystretch}{1.2}
\resizebox{\textwidth}{!}{
\begin{tabular}{ll|cc|ccc|c}
\toprule
\multirow{2}{*}{\textbf{Modality}} & \multirow{2}{*}{\textbf{Model}} & \multicolumn{2}{c|}{\textbf{\makecell{L1. Min/Max \\ Grounding}}} & \textbf{\makecell{L2. Local Pattern \\ Differentiation}} & \textbf{\makecell{L2. Global Pattern \\ Differentiation}} &  \textbf{\makecell{L2. Numerical \\ Perception}} & \textbf{\makecell{L3. Semantic \\ Understanding}} \\
\cmidrule{3-8}
& & Acc\% & half-Acc\% & Acc\% & Acc\% & Acc\% & Acc\% \\
\midrule
\multirow{2}{*}{Text (w/o index)} 
 & GPT-4o & 6.0 & 23.0 & 53.0 & 94.0 & \underline{73.0} & 67.0 \\
 & Qwen3-8B & 33.0 & 67.0 & 59.0 & 66.0 & 36.0 & 54.0 \\
\midrule
\multirow{2}{*}{Text (w/ index)} 
 & GPT-4o & 23.0 & 66.0 & 55.0 & \underline{98.0} & 62.0 & 62.0 \\
 & Qwen3-8B & \textbf{60.0} & \textbf{80.0} & 55.0 & 89.0 & 64.0 & 66.0 \\
\midrule
\multirow{2}{*}{Vision} 
 & GPT-4o & - & - & 55.0 & 93.0 & 58.0 & 61.0 \\
 & Qwen3-VL-8B & - & - & 65.0 & 88.0 & 47.0 & 35.0 \\
\midrule
Time Series & ChatTS & \underline{37.0} & 54.0 & 67.0 & 88.0 & 63.0 & \underline{70.0} \\
\midrule
\multirow{2}{*}{\makecell{Vision+Text \\ (w/o index)}}
 & GPT-4o & 8.0 & 33.0 & \textbf{69.0} & 97.0 & 68.0 & \textbf{71.0} \\
 & Qwen3-VL-8B & 5.0 & 29.0 & \underline{68.0} & 87.0 & 51.0 & 48.0 \\
\midrule
\multirow{2}{*}{\makecell{Vision+Text \\ (w/ index)}}
 & GPT-4o & 29.0 & \underline{77.0} & 55.0 & \textbf{99.0} & 61.0 & 54.0 \\
 & Qwen3-VL-8B & 19.0 & 70.0 & \textbf{69.0} & 89.0 & \textbf{78.0} & 46.0 \\
\bottomrule
\end{tabular}
}
\caption{Evaluation results on key tasks in the \textsc{HiTSR}.}
\label{tab:baseline}
\end{table*}

\begin{table*}[htbp]
\centering
\setlength{\tabcolsep}{4pt}
\resizebox{\textwidth}{!}{
\renewcommand{\arraystretch}{1.2}
\begin{tabular}{c|c|cccccc}
\toprule
Type & Dataset & Data & Modalities & CoT & Train-Test & Level & Open source \\
\midrule
\multirow{16 }{*}{General}
& TSQA \cite{timemqa} & Real & Text & \ding{56} & $\checkmark$ & L2-L4 & $\checkmark$ \\
& TimeMCQ2 \cite{mcq2} & Syn. & Text & \ding{56} & \ding{56} & L3 & $\checkmark$ \\
& TimeSeriesExam \cite{timeseriesexam} & Syn. & Text & \ding{56} & \ding{56} & L2 & $\checkmark$ \\
& ChatTime \cite{chattime} & Syn.\&Real & Text & \ding{56} & $\checkmark$ & L2, L4 & $\checkmark$ \\
& ChatTS \cite{chatts} & Syn. & TS & \ding{56} & $\checkmark$ & L2-L3 & $\checkmark$ \\
& BEDTime \cite{bedtime} & Syn.\&Real & Text\&Vision & \ding{56} & \ding{56} & L2 & $\checkmark$ \\
& MMTS-Bench \cite{mmtsbench} & Syn.\&Real & Text & \ding{56} & \ding{56} & L2-L4 & $\checkmark$ \\
& CaTS-Bench \cite{catsbench} & Real & Text\&Vision & \ding{56} & $\checkmark$ & L2 & $\checkmark$ \\
& TRQA \cite{trqa} & Real & Text & \ding{56} & $\checkmark$ & L3 & $\checkmark$ \\
& TimeSense \cite{timesense} & Syn. & TS & \ding{56} & $\checkmark$ & L2-L3 & \ding{56} \\
& TSR-Suite \cite{timeomni} & Real & Text & $\checkmark$ & $\checkmark$ & L3-L4 & \ding{56} \\
& TimeMMD \cite{timemmd} & Syn.\&Real & TS & \ding{56} & $\checkmark$ & L4 & $\checkmark$ \\
& Chat-TS \cite{chat-ts} & Real & TS & \ding{56} & $\checkmark$ & L2-L3 & \ding{56} \\
& TSAIA \cite{tsaia} & Syn.\&Real & Text & \ding{56} & $\checkmark$ & L3-L4 & $\checkmark$ \\
& CiK \cite{cik} & Syn.\&Real & Text & \ding{56} & \ding{56} & L4 & $\checkmark$ \\
& MTBench \cite{mtbench} & Real & Text & \ding{56} & \ding{56} & L3-L4 & $\checkmark$ \\
\midrule
\multirow{4}{*}{\makecell{Domain \& \\ Task- \\ Specific}}
& RAT40k \cite{timera} & Real & Text\&Vision & $\checkmark$ & $\checkmark$ & L2-L3 & $\checkmark$ \\
& ITFormer \cite{itformer} & Syn. & TS & \ding{56} & $\checkmark$ & L3 & $\checkmark$ \\
& ECG-Grounding \cite{ecgrounding} & Real & TS\&Vision & $\checkmark$ & $\checkmark$ & L3 & $\checkmark$ \\
& SciTS \cite{scits} & Real & TS & \ding{56} & $\checkmark$ & L3-L4 & \ding{56} \\
\midrule
General & \textsc{HiTSR} & Syn.\&Real & Text\&Vision & $\checkmark$ & $\checkmark$ & L1-L3 & $\checkmark$ \\
\bottomrule
\end{tabular}
}
\caption{Dataset properties. ``Syn.'' denotes the synthetic time series data, and ``Train-Test'' represents whether the corresponding dataset provides train-test splits instead of as an evaluation benchmark.}
\label{tab:data_sum}

\end{table*}

\begin{table*}[tbp]
\centering
\setlength{\tabcolsep}{2pt}
\renewcommand{\arraystretch}{1.5}
\resizebox{\textwidth}{!}{
\begin{tabular}{c|c|cccccccc}
\toprule
\textbf{Type} & \textbf{Model} & \textbf{\makecell{Diagnosis \\ Accuracy}} & \textbf{\makecell{Analysis \\ Completeness}} & \textbf{\makecell{Analysis \\ Relevance}} & \textbf{\makecell{Lead \\ Assessment \\ Coverage}} & \textbf{\makecell{Lead \\ Assessment \\ Accuracy}} & \textbf{\makecell{ECG \\ Feature \\ Grounding}} & \textbf{\makecell{Evidence \\Based \\ Reasoning}} & \textbf{\makecell{Clinical \\ Diagnostic \\ Fidelity}} \\
\midrule
\multirow{5}{*}{ID}  & PULSE         & 81.14 & 2.37 & 2.39 & 7.11 & 2.95 & 50.18 & 52.40 & 51.63 \\
                     & GEM (LLaVA)    & \textbf{87.24} & 4.41 & \underline{5.01} & \underline{71.07} & \underline{46.44} & \textbf{75.48} & \textbf{75.09} & \textbf{75.28} \\
                     & GEM (PULSE)    & \underline{86.49} & \underline{4.43} & 4.91 & 69.80 & 45.33 & \underline{74.95} &  \underline{74.70} & \underline{74.87} \\
                     & Qwen3-VL-8B    & 60.85 & 4.01 & 3.97 & 69.28 & 50.10 & 69.89 & 63.82 & 69.10 \\
                     \cmidrule(lr){2-10}
                     & \textsc{LLaTiSA} & 62.77 & \textbf{5.64} & \textbf{7.14} & \textbf{84.03} & \textbf{53.03} & 71.41 & 71.20 & 70.10 \\
\midrule
\multirow{5}{*}{OOD} & PULSE         & 59.24 & 2.20 & 2.06 & 11.20 & 6.27 & 52.52 & 55.48 & 53.85 \\
                     & GEM (LLaVA)    & \underline{73.53} & \textbf{4.19} & \underline{2.96} & \textbf{79.54} & \underline{49.01} & \underline{74.48} & \underline{74.61} & \underline{73.84} \\
                     & GEM (PULSE)    & \textbf{73.59} & \textbf{4.19} & \textbf{3.00} & \underline{78.86} & 47.96 & \textbf{74.97} & \textbf{75.41} & \textbf{74.24} \\
                     & Qwen3-VL-8B    & 59.01 & 3.22 & 1.82 & 56.41 & 38.12 & 46.43 & 63.81 & 49.89 \\
                     \cmidrule(lr){2-10}
                     & \textsc{LLaTiSA} & 62.22 & \underline{3.62} & 2.48 & 66.50 & \textbf{49.20} & 66.19 & 66.61 & 64.91 \\
\bottomrule
\end{tabular}
}
\caption{Full results on the L3 ECG-Grounding task. The best result is \textbf{bolded} and the second best is in \underline{underlined}.}
\label{tab:full_ecg}
\end{table*}

\begin{table*}[t]
\centering
\begin{tabularx}{\textwidth}{@{}>{\bfseries}p{1.8cm}X@{}}
\toprule
\multicolumn{2}{@{}l@{}}{\textbf{Model-Case1}}\\
\midrule

\multicolumn{2}{@{}p{\textwidth}@{}}{%
\textbf{Question: }Carefully analyze the provided time series plot in the image and choose the single best option that most accurately describes the pattern depicted.
Follow these rules strictly: (1) Look at the image before deciding.
(2) Only output the chosen option, highlighted as A, B, C, or D;
(3) Avoid adding extra text or explanations.
}\\[2mm]

\multicolumn{2}{@{}c@{}}{%
  \includegraphics[width=\textwidth]{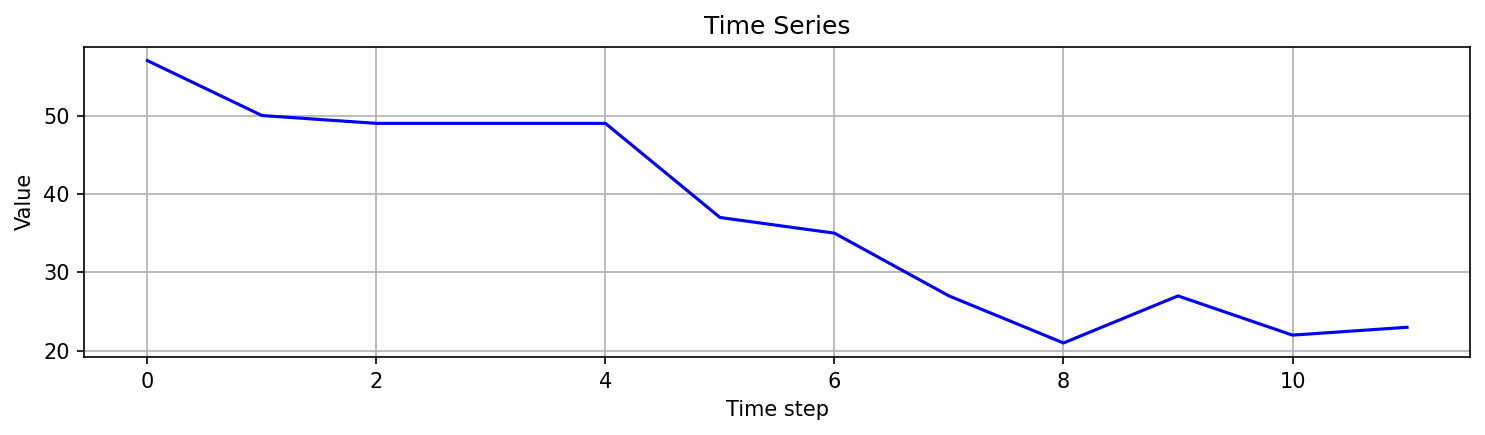}%
}\\[2mm]

\multicolumn{2}{@{}l@{}}{\textbf{Options:}}\\[-1mm]
\multicolumn{2}{@{}l@{}}{%
  \setlength{\tabcolsep}{4pt}%
  \renewcommand{\arraystretch}{1.05}%
  \begin{tabularx}{\textwidth}{@{}>{\bfseries}p{1.6em}X@{}}
    A. & Peaks towards the end. \\
    \textcolor{blue}{B.} & \textcolor{blue}{Ends at same value as beginning.} \\
    C. & Highest point at the end. \\
    \textcolor{red}{D.} & \textcolor{red}{Decreases smoothly in the beginning.} \\
    
  \end{tabularx}%
}\\[1mm]

\multicolumn{2}{@{}l@{}}{\bfseries Answer: B}\\
\bottomrule
\end{tabularx}
\caption{Bad cases generated by GPT-4o. Options in \textcolor{red}{red} denote the correct answer, and the \textcolor{blue}{blue} one is the wrong option chosen by the model.}
\label{tab:model_bad_case1}
\end{table*}

\begin{table*}[t]
\centering
\begin{tabularx}{\textwidth}{@{}>{\bfseries}p{1.8cm}X@{}}
\toprule
\multicolumn{2}{@{}l@{}}{\textbf{Model-Case2}}\\
\midrule

\multicolumn{2}{@{}p{\textwidth}@{}}{%
\textbf{Question: }Carefully analyze the provided time series plot in the image and choose the single best option that most accurately describes the pattern depicted.
Follow these rules strictly: (1) Look at the image before deciding.
(2) Only output the chosen option, highlighted as A, B, C, or D;
(3) Avoid adding extra text or explanations.
}\\[2mm]
 
\multicolumn{2}{@{}c@{}}{%
  \includegraphics[width=\textwidth]{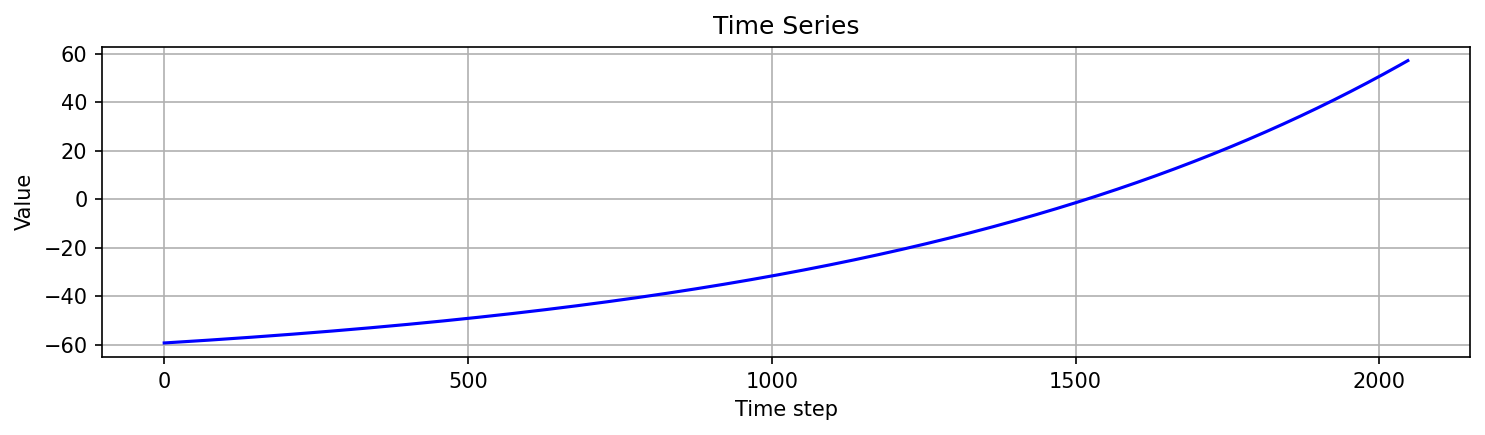}%
}\\[2mm]

\multicolumn{2}{@{}l@{}}{\textbf{Options:}}\\[-1mm]
\multicolumn{2}{@{}l@{}}{%
  \setlength{\tabcolsep}{4pt}%
  \renewcommand{\arraystretch}{1.05}%
  \begin{tabularx}{\textwidth}{@{}>{\bfseries}p{1.6em}X@{}}
    A. & A periodic pattern that is consistently repeated at regular intervals, similar to a square wave, is presented by the signal. Significant step changes punctuate the signal. \\
    \textcolor{red}{B.} & \textcolor{red}{The course is marked by exponential amplification, resulting in a sharp spike at the end.} \\
    \textcolor{blue}{C.} & \textcolor{blue}{Like an inverted sigmoid curve that forms a mirrored S-shape, the trend shows a monotonically decreasing pattern, with its rate of decrease gradually slowing down, eventually converging to a certain value. For the most part, the signal is smooth with slight disturbances.} \\
    D. & The trend is straight in its growth. Usually, the signal is smooth with minor spikes. \\
    
  \end{tabularx}%
}\\[1mm]

\multicolumn{2}{@{}l@{}}{\bfseries Answer: B}\\
\bottomrule
\end{tabularx}
\caption{Bad cases generated by GPT-4o. Options in \textcolor{red}{red} denote the correct answer, and the \textcolor{blue}{blue} one is the wrong option chosen by the model.}
\label{tab:model_bad_case2}
\end{table*}

\begin{table*}[t]
\centering
\begin{tabularx}{\textwidth}{@{}>{\bfseries}p{1.8cm}X@{}}
\toprule
\multicolumn{2}{@{}l@{}}{\textbf{Data-Case1: Multiple Correct Options}}\\
\midrule

\multicolumn{2}{@{}p{\textwidth}@{}}{%
\textbf{Question: }Carefully analyze the provided time series plot in the image and choose the single best option that most accurately describes the pattern depicted.
Follow these rules strictly: (1) Look at the image before deciding.
(2) Only output the chosen option, highlighted as A, B, C, or D;
(3) Avoid adding extra text or explanations.
}\\[2mm]

\multicolumn{2}{@{}c@{}}{%
  \includegraphics[width=\textwidth]{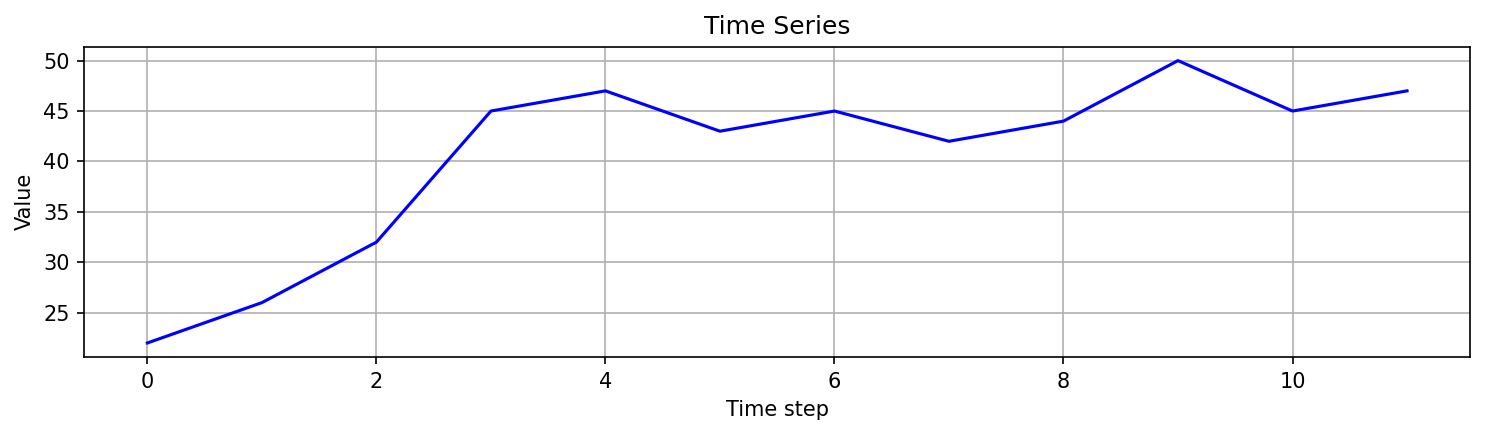}%
}\\[2mm]

\multicolumn{2}{@{}l@{}}{\textbf{Options:}}\\[-1mm]
\multicolumn{2}{@{}l@{}}{%
  \setlength{\tabcolsep}{4pt}%
  \renewcommand{\arraystretch}{1.05}%
  \begin{tabularx}{\textwidth}{@{}>{\bfseries}p{1.6em}X@{}}
    \textcolor{red}{A.} & \textcolor{red}{endpoint is higher than the beginning.} \\
    \textcolor{red}{B.} & \textcolor{red}{mostly flat in the middle.} \\
    C. & maximum value at the end. \\
    D. & big climb towards the end. \\
    
  \end{tabularx}%
}\\[1mm]

\multicolumn{2}{@{}l@{}}{\bfseries Answer: B}\\
\bottomrule
\end{tabularx}
\caption{Ambiguous questions observed from existing datasets. Options in \textcolor{red}{red} denote the correct answer.}
\label{tab:data_bad_case1}
\end{table*}

\begin{table*}[t]
\centering
\begin{tabularx}{\textwidth}{@{}>{\bfseries}p{1.8cm}X@{}}
\toprule
\multicolumn{2}{@{}l@{}}{\textbf{Data-Case2: The unclear definition of Location Terms}}\\
\midrule

\multicolumn{2}{@{}p{\textwidth}@{}}{%
\textbf{Question: }Carefully analyze the provided time series plot in the image and choose the single best option that most accurately describes the pattern depicted.
Follow these rules strictly: (1) Look at the image before deciding.
(2) Only output the chosen option, highlighted as A, B, C, or D;
(3) Avoid adding extra text or explanations.
}\\[2mm]

\multicolumn{2}{@{}c@{}}{%
  \includegraphics[width=\textwidth]{Figs/dataset_bad_case1.png}%
}\\[2mm]
\multicolumn{2}{@{}l@{}}{\textbf{Options:}}\\[-1mm]
\multicolumn{2}{@{}l@{}}{%
  \setlength{\tabcolsep}{4pt}%
  \renewcommand{\arraystretch}{1.05}%
  \begin{tabularx}{\textwidth}{@{}>{\bfseries}p{1.6em}X@{}}
    \textcolor{red}{A.} & \textcolor{red}{maximum value at the end.} \\
    \textcolor{red}{B.} & \textcolor{red}{dips at the end.} \\
    \textcolor{blue}{C.} & \textcolor{blue}{sharp decline near the end.} \\
    D. & ends lower than it started. \\
    
  \end{tabularx}%
}\\[1mm]

\multicolumn{2}{@{}l@{}}{\bfseries Answer: C}\\
\bottomrule
\end{tabularx}
\caption{A question sample that lacks a clear definition of location terms. Options in \textcolor{red}{red} denote are possibly correct, depending on the definition of ``end''. The option marked in \textcolor{blue}{blue} is the wrong option chosen by GPT-4o, which is the ``guessing the boundary of volatility'' problem indicated by \cite{timeomni}.}
\label{tab:data_bad_case2}
\end{table*}

\begin{table*}[t]
\centering
\begin{tabularx}{\textwidth}{@{}>{\bfseries}p{1.8cm}X@{}}
\toprule
\multicolumn{2}{@{}l@{}}{\textbf{Data-Case3: Lack of Information}}\\
\midrule

\multicolumn{2}{@{}p{\textwidth}@{}}{%
\textbf{Question: }Classify the given time series into one of the categories below.

Respond ONLY with the letter of the correct choice (A, B).

Choices:

A: normal process

B: abnormal process
}\\[2mm]

\multicolumn{2}{@{}c@{}}{%
  \includegraphics[width=\textwidth]{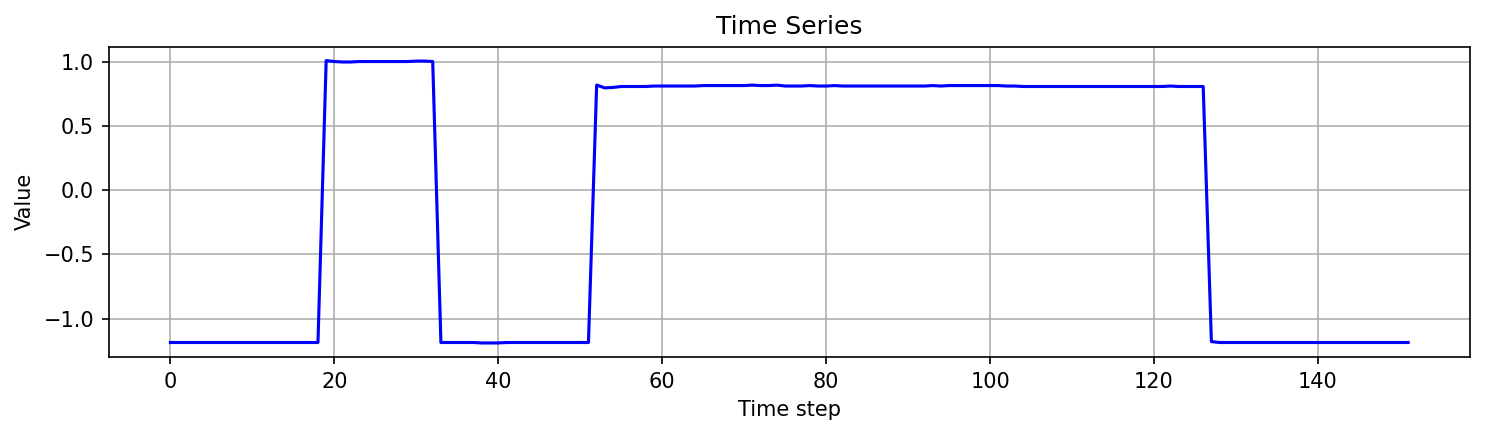}%
}\\[2mm]

\multicolumn{2}{@{}l@{}}{\bfseries Answer: B}\\
\bottomrule
\end{tabularx}
\caption{A question sample that lacks sufficient information, i.e., does not introduce the definition of ``normal process''.}
\label{tab:data_bad_case3}
\end{table*}

\begin{table*}[t]
\centering
\begin{tabularx}{\textwidth}{@{}>{\bfseries}p{1.8cm}X@{}}
\toprule
\multicolumn{2}{@{}l@{}}{\textbf{L1-} \emph{Numerical Read-out}: Min/Max Grounding}\\
\midrule

\multicolumn{2}{@{}p{\textwidth}@{}}{%
\textbf{Question: }Given the time series: <ts>, find the maximum and minimum values in the time series and report their first occurrence indices (if max/min repeats, use the earliest index). Provide the exact values and state which appears first. The first index is 0.

You MUST exactly follow the output format as:

<max>[index, value]</max>

<min>[index, value]</min>

The max value appears first.

OR

The min value appears first.
}\\[2mm]

\multicolumn{2}{@{}c@{}}{%
  \includegraphics[width=\textwidth]{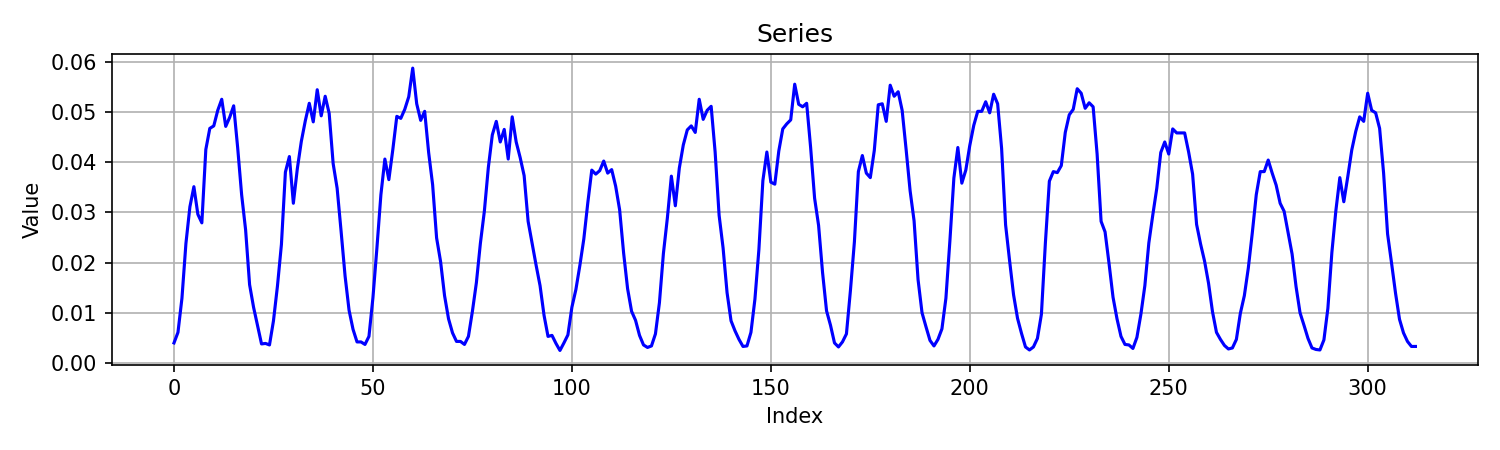}%
}\\[2mm]


\multicolumn{2}{@{}p{\textwidth}@{}}{%
\textbf{Answer:}\par
<max>[60, 0.0587]</max>\par
<min>[97, 0.0025]</min>\par
The max value appears first.
}\\
\bottomrule
\end{tabularx}
\caption{Illustration of ``Min/Max Grounding'' task. ``<ts>'' denotes various input types of time series, e.g., numerical tables, textual arrays, or specific time series tokens.}
\label{tab:l1_example_1}
\end{table*}

\begin{table*}[t]
\centering
\begin{tabularx}{\textwidth}{@{}>{\bfseries}p{1.8cm}X@{}}
\toprule
\multicolumn{2}{@{}l@{}}{\textbf{L1-} \emph{Numerical Read-out}: Start/End Comparsion} \\
\midrule

\multicolumn{2}{@{}p{\textwidth}@{}}{%
\textbf{Question: }Given the time series: <ts>, compare the value at the start of the time series (index 0) and the value at the end (last index). Report both [index, value] pairs and then state the comparison result.

Output format (MUST follow exactly):

<start>[index, value]</start>

<end>[index, value]</end>

The start value is larger than the value at the end.

OR

The start value is smaller than the value at the end.

OR

The start value is equal to the value at the end.
}\\[2mm]

\multicolumn{2}{@{}c@{}}{%
  \includegraphics[width=\textwidth]{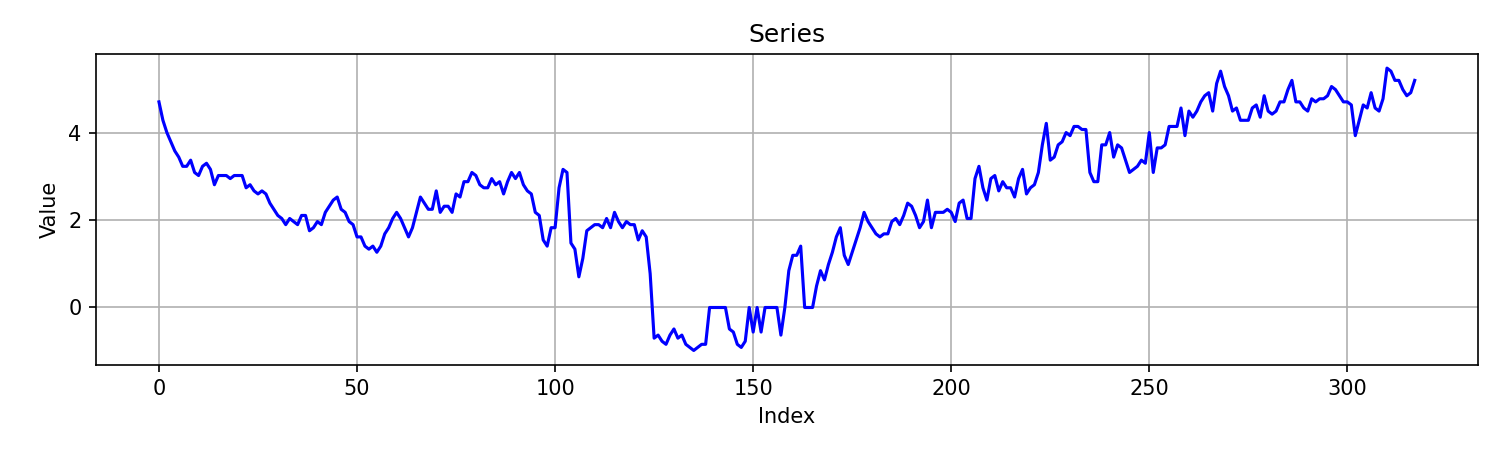}%
}\\[2mm]


\multicolumn{2}{@{}p{\textwidth}@{}}{%
\textbf{Answer:}\par
<start>[0, 4.713]</start>\par
<end>[317, 5.206]</end>\par
The start value is smaller than the value at the end.
}\\
\bottomrule
\end{tabularx}
\caption{Illustration of ``Start/End Comparsion'' task. ``<ts>'' denotes various input types of time series, e.g., numerical tables, textual arrays, or specific time series tokens.}
\label{tab:l1_example_2}
\end{table*}

\begin{table*}[t]
\centering
\begin{tabularx}{\textwidth}{@{}>{\bfseries}p{1.8cm}X@{}}
\toprule
\multicolumn{2}{@{}l@{}}{\textbf{L1-} \emph{Numerical Read-out}: Mulri-series Comparsion}\\
\midrule

\multicolumn{2}{@{}p{\textwidth}@{}}{%
\textbf{Question: }Given the time series: <ts>, answer the question based on the provided multiple time series.

You MUST exactly follow the output format as:

<answer>[index, value]</answer>

<series>X</series>, 

where X refers to the index of the time series.

Question: Compare the minimum values of the multiple given time series. Which series has the lowest minimum value, and what is that specific value?
}\\[2mm]

\multicolumn{2}{@{}c@{}}{%
  \includegraphics[width=\textwidth]{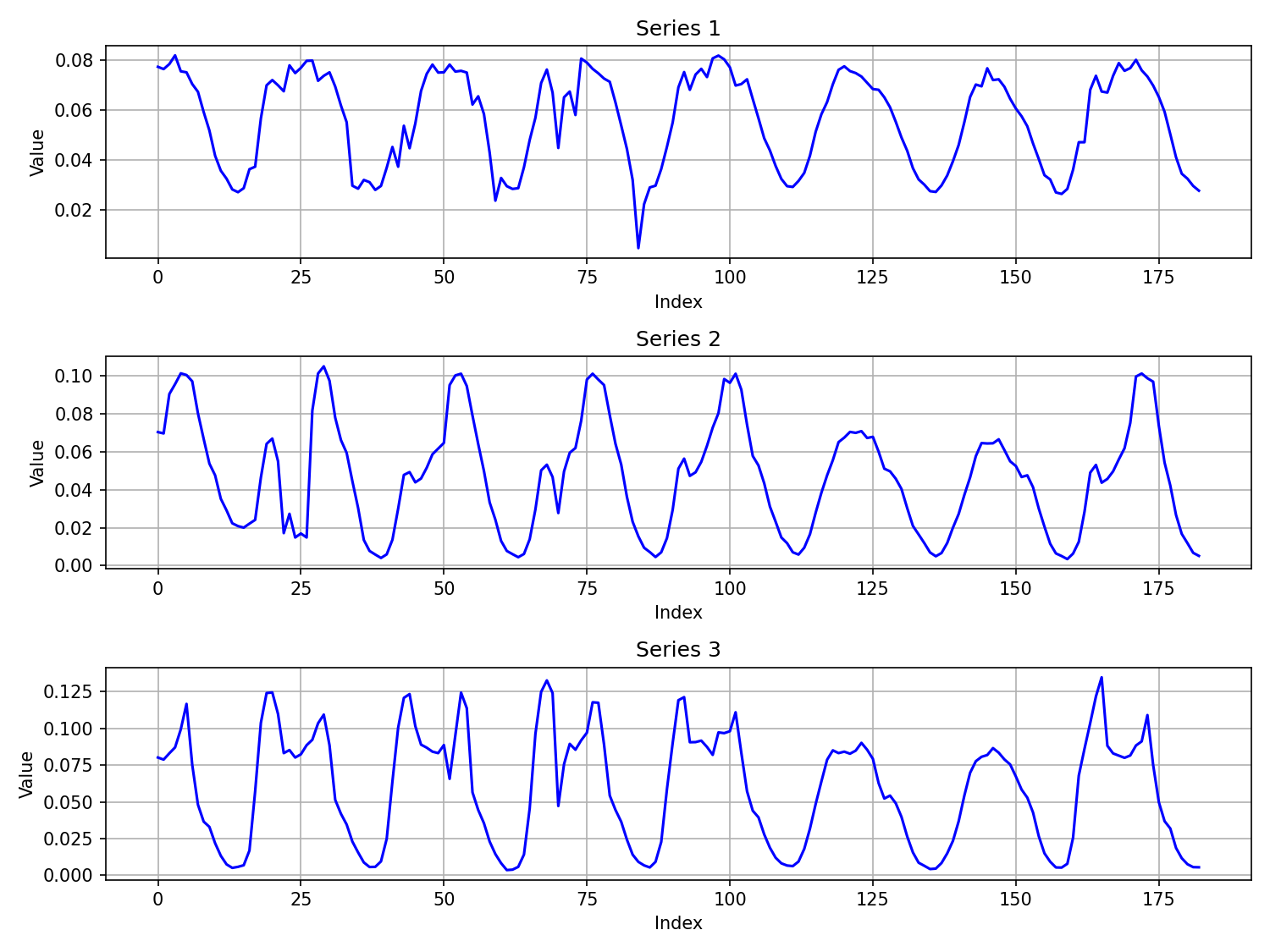}%
}\\[2mm]


\multicolumn{2}{@{}p{\textwidth}@{}}{%
\textbf{Answer:}\par
<answer>[159, 0.0034]</answer>\par

<series>2</series>\par

Time Series 2 has the lowest minimum value among all the series. Its exact minimum value is 0.0034.
}\\
\bottomrule
\end{tabularx}
\caption{Illustration of ``Multi-series Comparsion'' task. ``<ts>'' denotes various input types of time series that represent numerical information, e.g., numerical tables, textual arrays, or specific time series tokens.}
\label{tab:l1_example_3}
\end{table*}

\begin{table*}[t]
\centering
\begin{tabularx}{\textwidth}{@{}>{\bfseries}p{1.8cm}X@{}}
\toprule
\multicolumn{2}{@{}l@{}}{\textbf{L1-} \emph{Numerical Read-out}: Subseries Localization}\\
\midrule

\multicolumn{2}{@{}p{\textwidth}@{}}{%
\textbf{Question: }Given the time series: <ts>, answer the question based on the provided multiple time series.

You MUST exactly follow the output format as:

<answer>[v1, v2, ..., vk]</answer>

Recover the values of series 4 from index 201 to index 217 (inclusive). Return the values as a list in the exact order.
}\\[2mm]

\multicolumn{2}{@{}c@{}}{%
  \includegraphics[width=\textwidth]{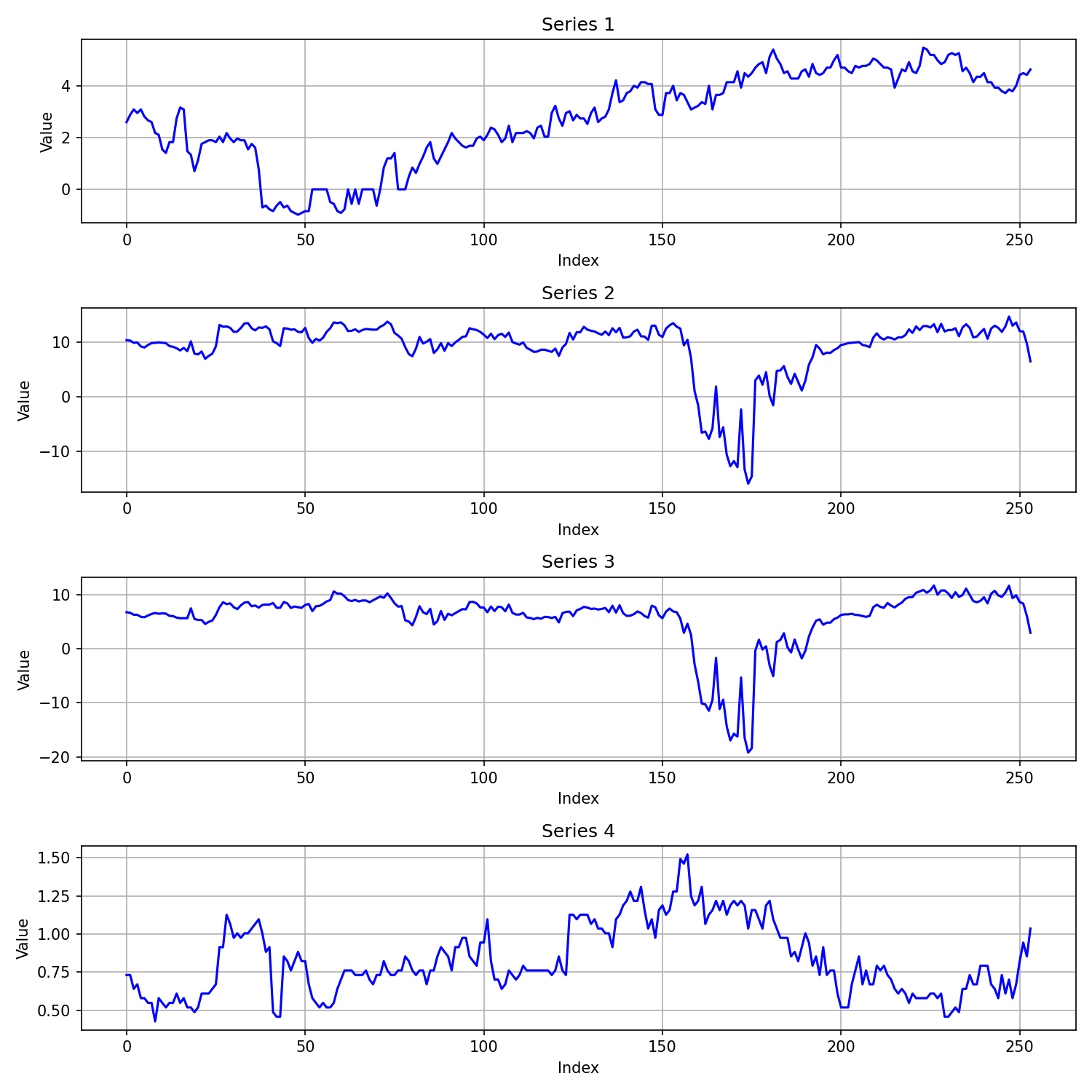}%
}\\[2mm]


\multicolumn{2}{@{}p{\textwidth}@{}}{%
\textbf{Answer:}\par
<answer>[0.518, 0.518, 0.67, 0.761, 0.853, 0.67, 0.761, 0.67, 0.67, 0.792, 0.761, 0.792, 0.731, 0.701, 0.64, 0.609, 0.64]</answer>

}\\
\bottomrule
\end{tabularx}
\caption{Illustration of ``Subseries Localization'' task. ``<ts>'' denotes various input types of time series that represent numerical information, e.g., numerical tables, textual arrays, or specific time series tokens.}
\label{tab:l1_example_4}
\end{table*}

\begin{table*}[t]
\centering
\begin{tabularx}{\textwidth}{@{}>{\bfseries}p{1.8cm}X@{}}
\toprule
\multicolumn{2}{@{}l@{}}{\textbf{L2-}\emph{Pattern Perception}: Local Pattern Differentiation}\\
\midrule

\multicolumn{2}{@{}p{\textwidth}@{}}{%
\textbf{Question: } Given the time series: <ts>, carefully analyze each option, then choose the single best option that most accurately describes the pattern.

Options:

A: The approximate minimum value of the series is about -24059.

B: The time series exhibits a square-wave periodic fluctuation, with an approximate minimum period length of 23.1.

C: The value at the end of the series is higher than at the start.

D: The time series contains 0 overall trend turning points. Turning points refer to changes in the overall trend direction.

}\\[2mm]

\multicolumn{2}{@{}c@{}}{%
  \includegraphics[width=\textwidth]{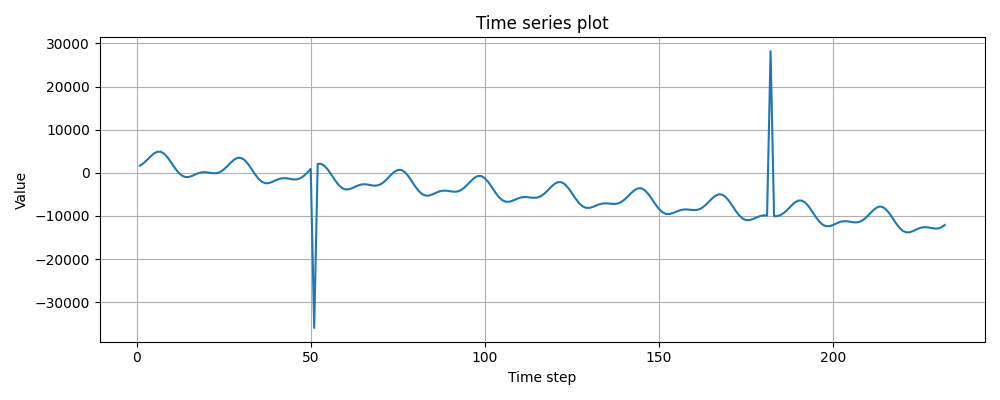}%
}\\[2mm]


\multicolumn{2}{@{}p{\textwidth}@{}}{%
\textbf{Answer:}
Step-by-step analysis of each option:\par

A: The series has a deep negative spike around time $\sim$50 that reaches close to -34,000 (near the -35,000 axis tick). -24,059 is far above this minimum, so this is inaccurate.\par

B: The plot is not a square wave; it looks more like a smooth oscillation with noise and two sharp outliers. Claiming a square-wave periodic fluctuation and a specific period (23.1) does not match the visual pattern, so this is inaccurate.\par

C: The series starts near a positive value (a few thousand) and ends well below zero (around -10,000 to -12,000). The end value is lower than the start, so this is false.\par

D: The overall trend drifts downward from start to end without a sustained reversal in direction; the large spikes are brief anomalies, not changes in long-term trend. Thus, the number of overall trend turning points is effectively 0.\par

The correct answer is D.
}\\
\bottomrule
\end{tabularx}
\caption{Illustration of ``Local Pattern Differentiation'' task in L2. ``<ts>'' denotes various input types of time series that represent numerical information, e.g., numerical tables, textual arrays, or specific time series tokens.}
\label{tab:l2_example_1}
\end{table*}

\begin{table*}[t]
\centering
\begin{tabularx}{\textwidth}{@{}>{\bfseries}p{1.8cm}X@{}}
\toprule
\multicolumn{2}{@{}l@{}}{\textbf{L2-}\emph{Pattern Perception}: Global Pattern Differentiation}\\
\midrule

\multicolumn{2}{@{}p{\textwidth}@{}}{%
\textbf{Question: } Given the time series: <ts>, carefully analyze each option, then choose the single best option that most accurately describes the pattern.

Options:

A. This time series steadily decreases from about 91 at index 0 (its maximum) to about 13 at index 162 (its minimum), with a smooth low-volatility path and no apparent turning points or seasonality.

B. This time series gently declines from index 0 around −200 to its minimum near -500 at index 80, then steadily climbs with moderate volatility to finish at its maximum around 260 at index 120, showing no apparent seasonality.

C. Starting near 200, the series spikes to around 2000 at index 11, then plunges below -1200 by around index 17 and continues as a steady, low-volatility sine-like oscillation with a roughly 26-point period around -1100, reaching a valley near $-$1380 at index 194 and ending close to -1000 without significant turning points.

D. This time series gently declines from index 390 around -42 to its minimum near -979 at index 615, then steadily climbs with moderate volatility to finish at its maximum around 554 at index 634, showing no apparent seasonality or major turning points.

}\\[2mm]

\multicolumn{2}{@{}c@{}}{%
  \includegraphics[width=\textwidth]{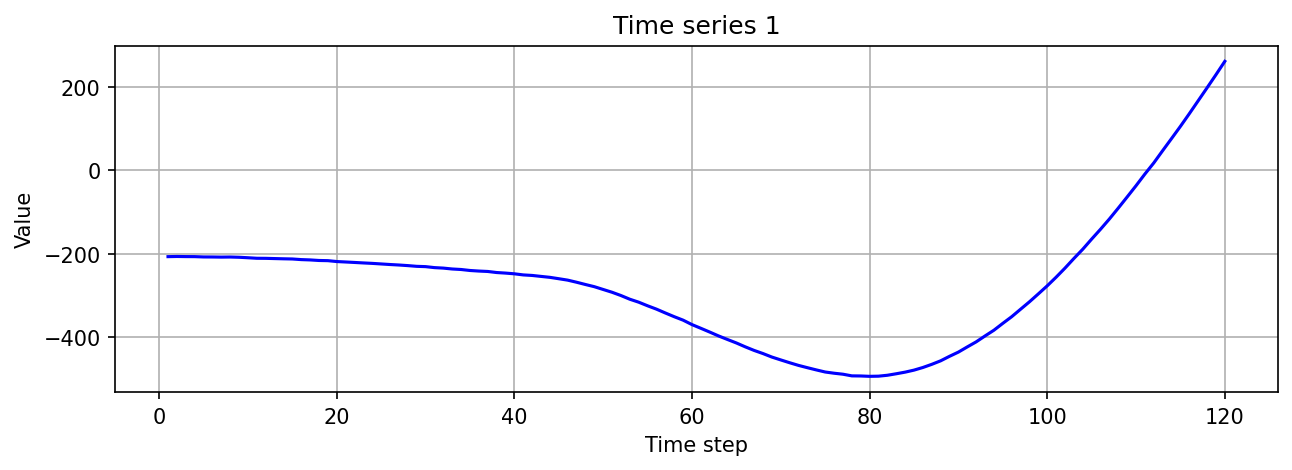}%
}\\[2mm]


\multicolumn{2}{@{}p{\textwidth}@{}}{%
\textbf{Answer:}
Step-by-step analysis:\par

A. Describes a steady decrease from about 91 at index 0 to about 13 at index 162, with no turning points. The provided plot starts around -200, decreases to roughly -500 by $\sim$80, then rises to about 260 by 120. It is not monotonic and ends positively, so A does not match.\par

B. Describes a gentle decline from around −200 at index 0 to a minimum near -500 at index 80, then a steady climb with moderate volatility to finish at a maximum around 260 at index 120, with no seasonality. This matches the plot very well: start near $-$200, bottom around −500 at $\sim$80, and end around 260 at 120, with smooth behavior and no seasonality.\par

C. Describes a series starting near 200, spiking to $\sim$2000 at index 11, plunging below -1200 by $\sim$17, then continuing as a sine-like oscillation around -1100, with a valley near index 194 and ending near -1000. The plot shows none of these extreme values or oscillatory behavior, nor does it extend to index 194. Not a match.\par

D. Describes a series starting at index 390 around -42, falling to -979 at index 615, then climbing to 554 at index 634. The indices and values are completely different from the plot (which spans 0 to 120). Not a match.\par

The correct answer is B.
}\\
\bottomrule
\end{tabularx}
\caption{Illustration of ``Global Pattern Differentiation'' task in L2. ``<ts>'' denotes various input types of time series that represent numerical information, e.g., numerical tables, textual arrays, or specific time series tokens.}
\label{tab:l2_example_2}
\end{table*}

\begin{table*}[t]
\centering
\begin{tabularx}{\textwidth}{@{}>{\bfseries}p{1.8cm}X@{}}
\toprule
\multicolumn{2}{@{}l@{}}{\textbf{L2-}\emph{Pattern Perception}: Numerical Perception}\\
\midrule

\multicolumn{2}{@{}p{\textwidth}@{}}{%
\textbf{Question: } Given the time series: <ts>, carefully analyze each option, choose the option that correctly describes \textbf{BOTH} the numerical features and the overall shape.

\textbf{Constraint}: You must select the option with the most precise numerical value.

Options:

A. The starting value is 7.950, and the overall trend is increasing.

B. The minimum value is 3.295 at index 90, and the overall trend is increasing.

C. The minimum value is 3.016 at index 90, and the overall trend is increasing.

D. The minimum value is 3.016 at index 90, and the overall trend is decreasing.

}\\[2mm]

\multicolumn{2}{@{}c@{}}{%
  \includegraphics[width=\textwidth]{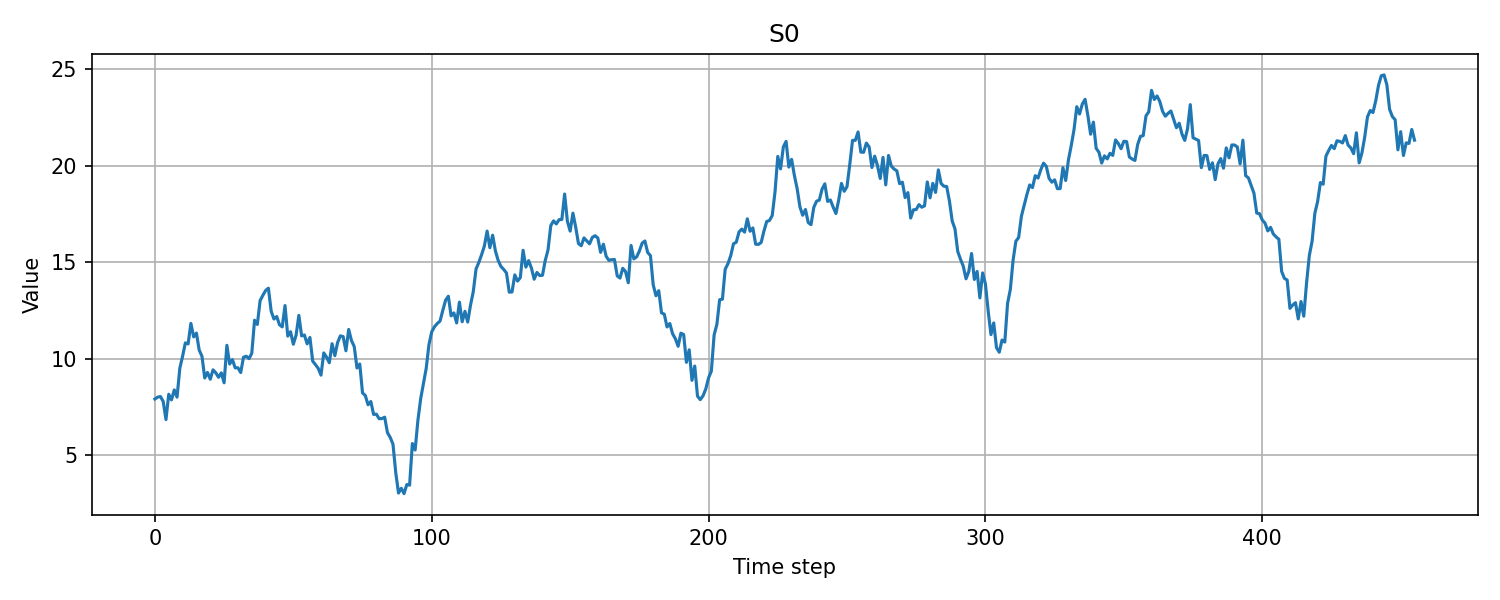}%
}\\[2mm]


\multicolumn{2}{@{}p{\textwidth}@{}}{%
\textbf{Answer:}\par

Option A: incorrect. Although the shape description matches, the numerical value 7.950 is imprecise. According to the numerical information, the correct value for min is 3.016. 

Option B: incorrect. Although the shape description matches, the numerical value 3.295 is imprecise. According to the numerical information, the correct value for min is 3.016. 

Option D: incorrect. While the numerical value is correct, the shape description `the overall trend is decreasing' contradicts the visual plot. The actual trend is that the overall trend is increasing. 

The correct answer is C.

}\\
\bottomrule
\end{tabularx}
\caption{Illustration of ``Numerical Perception'' task in L2. ``<ts>'' denotes various input types of time series that represent numerical information, e.g., numerical tables, textual arrays, or specific time series tokens.}
\label{tab:l2_example_3}
\end{table*}

\begin{table*}[t]
\centering
\begin{tabularx}{\textwidth}{@{}>{\bfseries}p{1.8cm}X@{}}
\toprule
\multicolumn{2}{@{}l@{}}{\textbf{L3-}\emph{Semantic Reasoning}}\\
\midrule

\multicolumn{2}{@{}p{\textwidth}@{}}{
\textbf{Question: } Given the time series: <ts>, considering the overall pattern, precise numerical values, and semantic logic, which of the following scenarios best fits the given time series? You need first to check the overall pattern, then verify the precise values, and finally think whether the series exactly matches the given background knowledge.

A: The time series shows a periodic daily cycle with peak power outputs consistently occurring every 12th observation, specifically at index 12 (324.45 kW) and index 24 (1350.66 kW), reflecting regular wind patterns in a Chinese wind farm.

B: There are multiple zero or near-zero power readings (e.g., index 1: 0.0 kW, index 33: -0.3 kW) that align with turbine shutdowns during low wind conditions, and the highest recorded power output is 1525.51 kW at index 46, which is physically plausible for a modern wind turbine.

C: The power output increases monotonically after index 30, rising from -0.3 kW to 618.11 kW by index 58, indicating a steady wind acceleration over 48 minutes, consistent with sudden gusts in arid regions of China.

D: The value at index 5 is 1047.14 kW, and it decreases to 1077.57 kW at index 6, which represents a typical fluctuation due to wind turbulence, and this behavior occurs every 5 observations, forming a repeating short-term pattern.

}\\[2mm]

\multicolumn{2}{@{}c@{}}{%
  \includegraphics[width=0.7\textwidth]{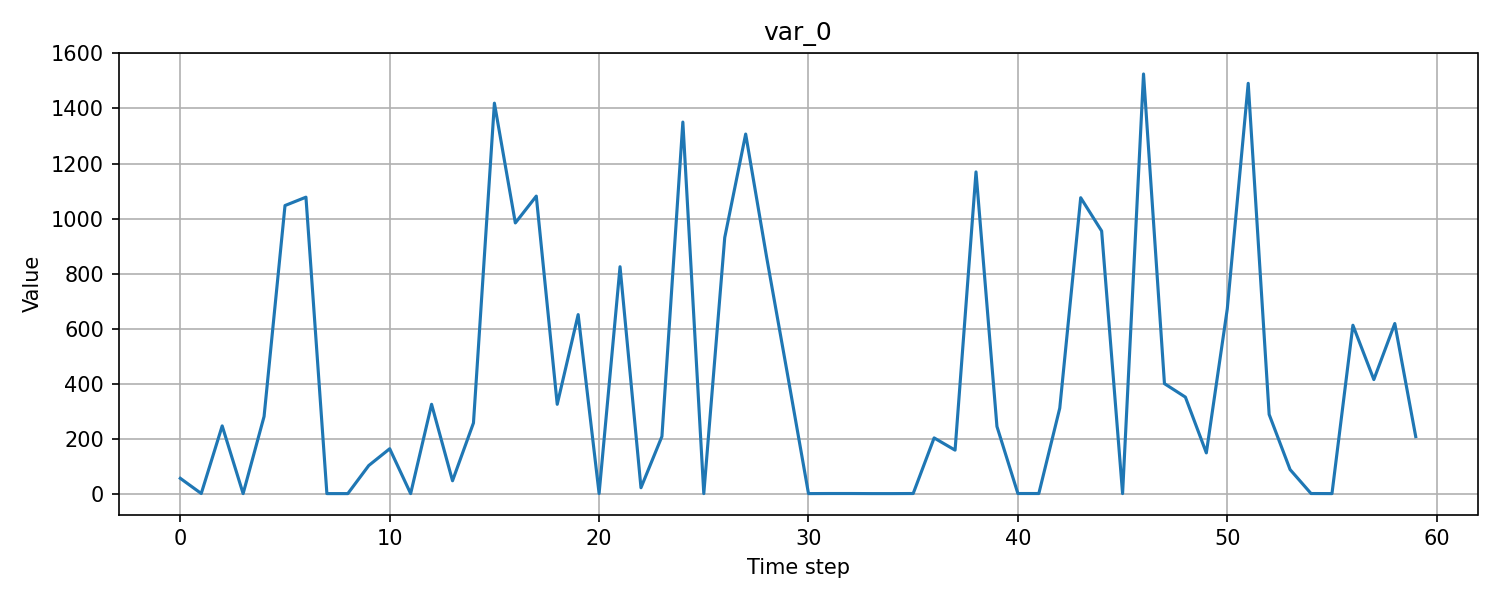}%
}\\[2mm]


\multicolumn{2}{@{}p{\textwidth}@{}}{
\textbf{Answer:}
Step-by-step analysis:\par

Option A: The claim of a periodic daily cycle every 12 observations is invalid because 12 observations correspond to 2 hours (12 * 10 minutes), not a full day; additionally, peak values do not consistently occur every 12 steps (e.g., index 15: 1419.46 kW, index 46: 1525.51 kW), and the pattern is irregular. This is a Pattern Error.\par

Option B: The statement correctly identifies multiple zero and near-zero values (e.g., index 1: 0.0, index 33: -0.3), which are plausible during turbine idling or low wind; the maximum value 1525.51 kW at index 46 is consistent with typical wind turbine capacity (1.5 MW class), and the context aligns with real-world operation in China. No errors detected.\par

Option C: The claim of monotonic increase after index 30 is false: from index 30 (-0.3) to 31 (0.0) to 32 (0.0) to 33 (-0.3), the series does not rise steadily; further, from index 34 onward, values fluctuate (e.g., index 35: 0.0, index 36: 202.17) but not monotonically. The assertion of a steady rise is contradicted by data. This is a Value Error.\par

Option D: The statement claims index 5 (1047.14) decreases to index 6 (1077.57), but 1077.57 is greater than 1047.14, so it is an increase, not a decrease. Additionally, no repeating 5-observation pattern is evident in the data. This contains a Value Error due to a correct comparison.

Therefore, after eliminating options with incorrect or unsupported interpretations, only \textbf{Option B} remains fully consistent with the data in terms of pattern, values, and semantics.\par

<answer>B</answer>

}\\
\bottomrule
\end{tabularx}
\caption{Illustration of ``Semantic Understanding'' task in L3. ``<ts>'' denotes various input types of time series that represent numerical information, e.g., numerical tables, textual arrays, or specific time series tokens.}
\label{tab:l3_example}
\end{table*}

\begin{table*}[t]
\centering
\begin{tabularx}{\textwidth}{@{}>{\bfseries}p{1.8cm}X@{}}
\toprule
\multicolumn{2}{@{}l@{}}{\textbf{Case: L1. Min/Max Grounding}}\\
\midrule

\multicolumn{2}{@{}p{\textwidth}@{}}{%
\textbf{Prompt: }<image><image>
You must refer to the second provided image to answer the question. The first image plots the visual trends. The second image is a \textbf{High-Density Numeric Grid} that provides precise values. 

\textbf{How to read the second image (Numerical Grid):}

1. \textbf{Layout}: The data is organized in a multi-column newspaper layout. Read the columns from \textbf{left to right}.

2. \textbf{Vertical Flow}: Within each major column, read the data vertically from \textbf{top to bottom}. When a column reaches the bottom, the sequence continues at the top of the next column to the right.

3. \textbf{Structure}: Inside each column block, the \textbf{leftmost sub-column labeled `T'} represents the time index. The subsequent sub-columns (e.g., Series) represent the values of different time series.

\textbf{Question:}
Find the maximum and minimum values in the time series and report their first occurrence indices (if max/min repeats, use the earliest index). Provide the exact values and state which appears first. The first index is 0.

You MUST exactly follow the output format as:

<max>[index, value]</max>
<min>[index, value]</min>

The max value appears first.

OR

The min value appears first.
}\\[2mm]

\multicolumn{2}{@{}c@{}}{%
    \centering
    \begin{minipage}{0.49\textwidth}
      \centering
      \includegraphics[width=\linewidth]{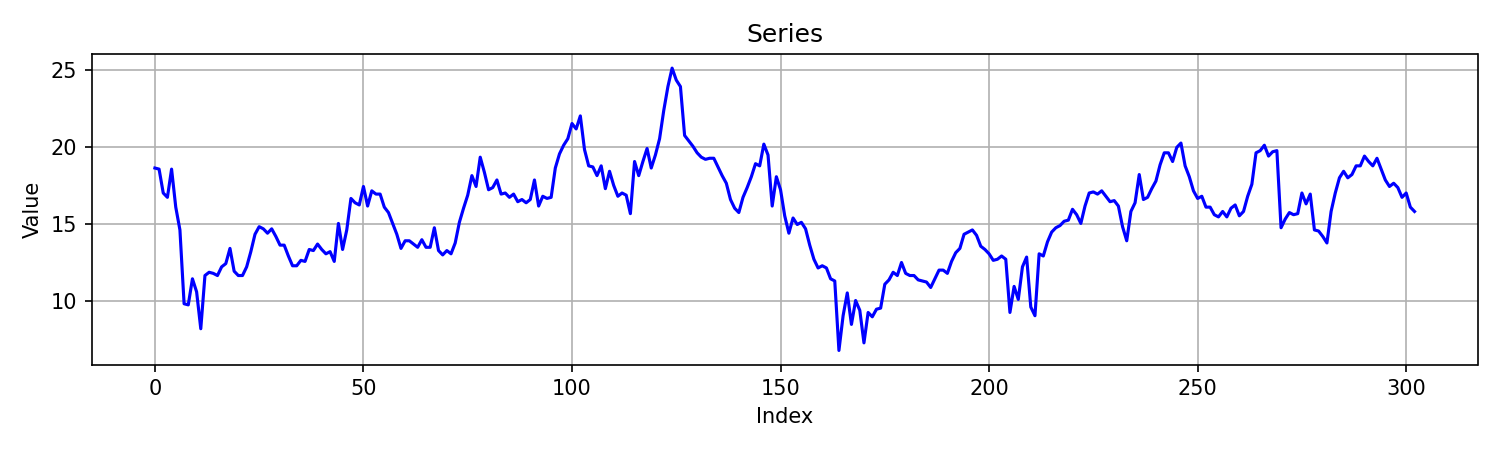}
    \end{minipage}\hfill
    \begin{minipage}{0.49\textwidth}
      \centering
      \includegraphics[width=\linewidth]{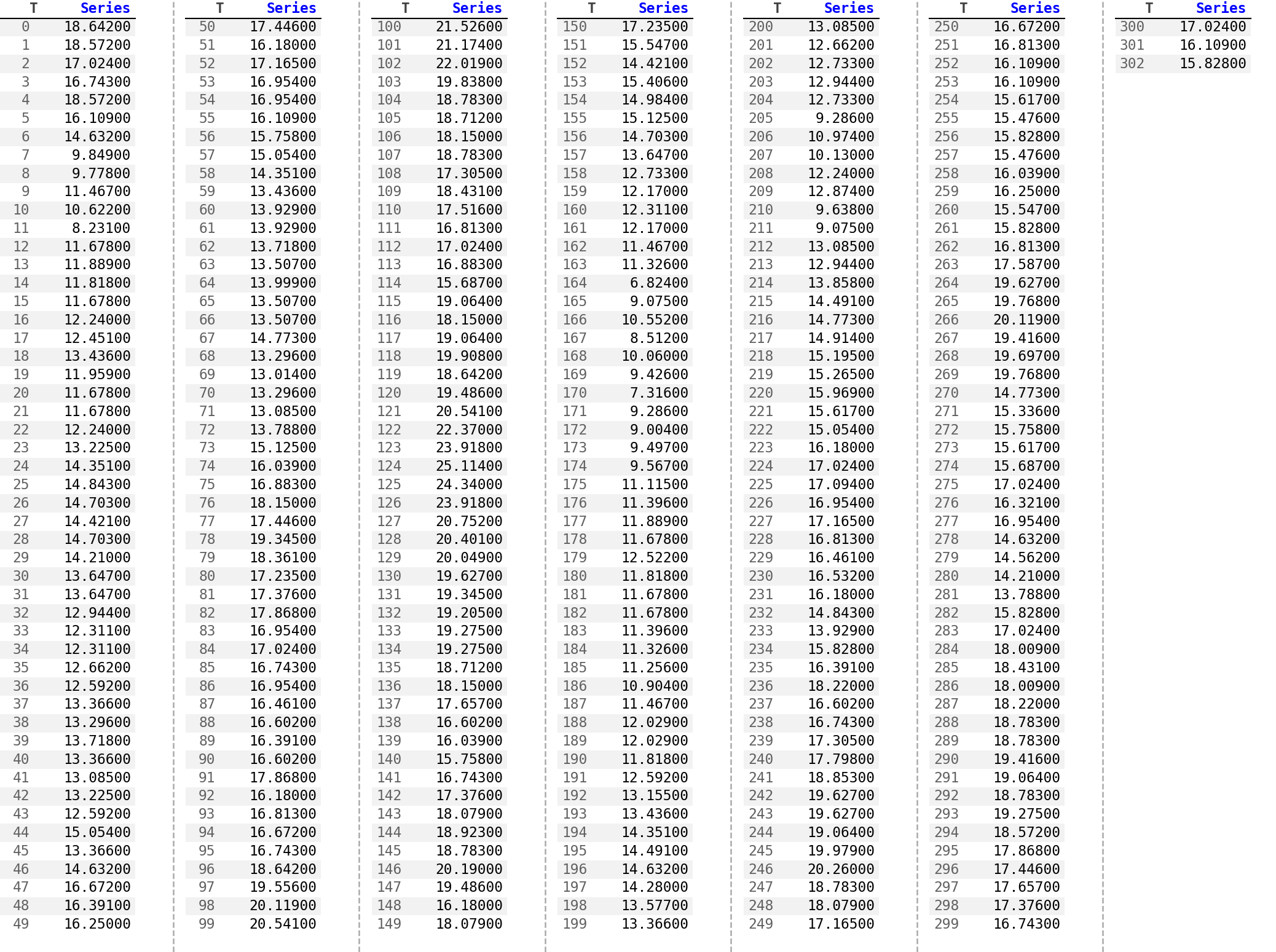}
    \end{minipage}\par
}\\[2mm]

\multicolumn{2}{@{}p{\textwidth}@{}}{%
\textbf{Answer: }

<max>[124, 25.11400]</max>

<min>[164, 6.82400]</min>

The max value appears first.} \\
\bottomrule
\end{tabularx}
\caption{``Min/Max Grounding'' OOD example with \textsc{LLaTiSA}'s response.}
\label{tab:llatisa_l1}
\end{table*}

\begin{table*}[t]
\centering
\begin{tabularx}{\textwidth}{@{}>{\bfseries}p{1.8cm}X@{}}
\toprule
\multicolumn{2}{@{}l@{}}{\textbf{Case: L2. Local Pattern Differentiation}}\\
\midrule
\multicolumn{2}{@{}p{\textwidth}@{}}{%
\textbf{Prompt: }<image><image>
You must refer to the second provided image to answer the question. The first image plots the visual trends. The second image is a \textbf{High-Density Numeric Grid} that provides precise values. 

\textbf{How to read the second image (Numerical Grid):}

1. \textbf{Layout}: The data is organized in a multi-column newspaper layout. Read the columns from \textbf{left to right}.

2. \textbf{Vertical Flow}: Within each major column, read the data vertically from \textbf{top to bottom}. When a column reaches the bottom, the sequence continues at the top of the next column to the right.

3. \textbf{Structure}: Inside each column block, the \textbf{leftmost sub-column labeled `T'} represents the time index. The subsequent sub-columns (e.g., Series) represent the values of different time series.

\textbf{Question:}

Carefully analyze the given time series description and choose the single best option that most accurately describes the pattern for the time series.

Options:

A: decreased in the middle

B: slow and steady decrease

C: ends with a drastic increase

D: begins with a shallow increase
}\\
\multicolumn{2}{@{}c@{}}{%
    \centering
    \begin{minipage}{0.49\textwidth}
      \centering
      \includegraphics[width=\linewidth]{Figs/llatisa_l1_1.png}
    \end{minipage}\hfill
    \begin{minipage}{0.49\textwidth}
      \centering
      \includegraphics[width=\linewidth]{Figs/llatisa_l1_2.png}
    \end{minipage}
}\\
\multicolumn{2}{@{}p{\textwidth}@{}}{%
\textbf{Answer: }

Option A: ``decreased in the middle'' — The series does dip in the middle, but the description is vague and doesn’t capture the later strong rebound and final rise. It’s partially true, but not the best summary.

Option B: ``slow and steady decrease'' — Incorrect. The series is not steadily decreasing; it has pronounced rises and falls, a trough around the middle, and a strong rise at the end.

Option C: ``ends with a drastic increase'' — Correct. After a low near the middle and a modest rise, the final segment jumps sharply from mid-50s to around 80, clearly a drastic increase.

Option D: ``begins with a shallow increase'' — The start does show a small rise (around 1–2 steps), but this is a minor detail and doesn’t describe the overall pattern as well as the strong final increase.

The correct answer is C.}\\
\bottomrule
\end{tabularx}
\caption{``Local Pattern Differentiation'' OOD example with \textsc{LLaTiSA}'s response. The question is from BEDTime \cite{bedtime}.}
\label{tab:llatisa_l2_local}
\end{table*}

\begin{table*}[t]
\centering
\begin{tabularx}{\textwidth}{@{}>{\bfseries}p{1.8cm}X@{}}
\toprule
\multicolumn{2}{@{}l@{}}{\textbf{Case: L2. Global Pattern Differentiation}} \\
\midrule

\multicolumn{2}{@{}p{\textwidth}@{}}{%
\textbf{Prompt: }<image><image>
You must refer to the second provided image to answer the question. The first image plots the visual trends. The second image is a \textbf{High-Density Numeric Grid} that provides precise values. 

\textbf{How to read the second image (Numerical Grid):}

1. \textbf{Layout}: The data is organized in a multi-column newspaper layout. Read the columns from \textbf{left to right}.

2. \textbf{Vertical Flow}: Within each major column, read the data vertically from \textbf{top to bottom}. When a column reaches the bottom, the sequence continues at the top of the next column to the right.

3. \textbf{Structure}: Inside each column block, the \textbf{leftmost sub-column labeled `T'} represents the time index. The subsequent sub-columns (e.g., Series) represent the values of different time series.

\textbf{Question:}

Given the provided time series plot with its value table, select the description from the four options that best corresponds to the time series without overthinking.

Options:

A) The data shows regular oscillatory behavior with consistent amplitude variations, featuring recurring peaks around 20-30 and baseline values near 1-2 throughout the sequence.

B) This time series exhibits cyclical patterns with periodic spikes reaching peaks around 15-20, followed by gradual declines to near-zero values before repeating the cycle.

C) This series demonstrates relatively stable fluctuations around 2-3 with occasional upward excursions, showing gradual trending behavior and moderate variability across the observation period.

D) The data maintains high baseline values around 87-94 with occasional step-like transitions and brief periods of lower values, exhibiting overall stability with discrete level changes.
}\\[2mm]

\multicolumn{2}{@{}c@{}}{%
    \centering
    \begin{minipage}{0.49\textwidth}
      \centering
      \includegraphics[width=\linewidth]{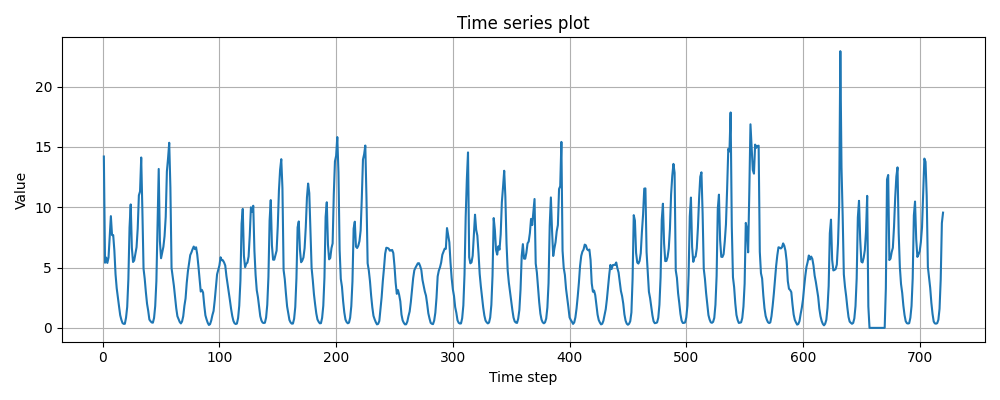}
    \end{minipage}\hfill
    \begin{minipage}{0.49\textwidth}
      \centering
      \includegraphics[width=\linewidth]{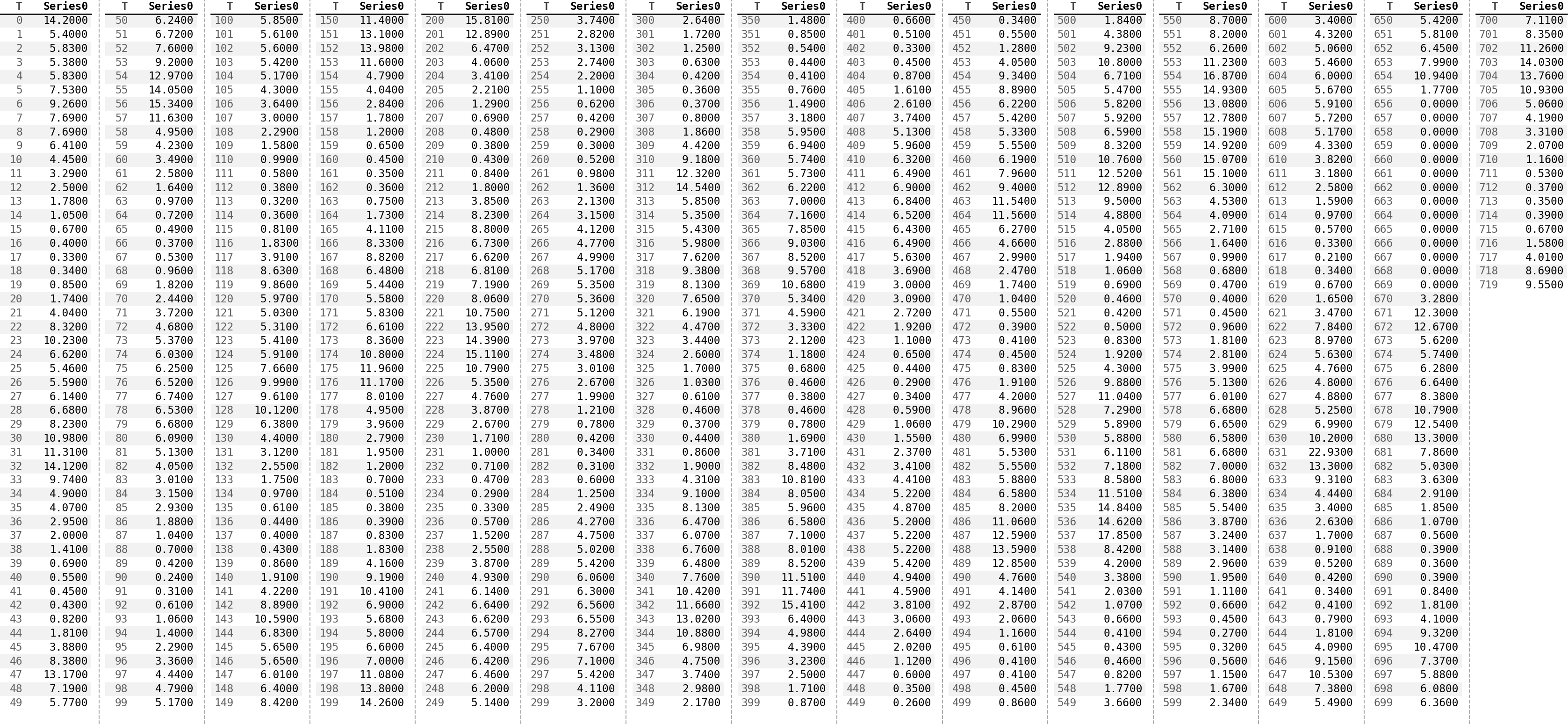}
    \end{minipage}\par
} \\

\multicolumn{2}{@{}p{\textwidth}@{}}{%
\textbf{Answer: }

Option A: Not a good match. While the series does oscillate, the peaks are not consistently in the 20–30 range (many are 10–15, and there’s a maximum around $\sim$23). The baseline is not persistently near 1–2; it sits closer to 3–6 with frequent oscillations. The “consistent amplitude variations” phrasing is also inconsistent with the plot’s changing peak heights.

Option B: Best match. The plot shows recurring cyclical behavior with spikes that often reach roughly 15 units (some cycles peak higher, around 18–22). After spikes, values frequently drop toward or briefly touch zero before the cycle repeats. This description aligns well with the observed oscillatory pattern and magnitude.

Option C: Not a match. The series is not centered around 2–3, nor are the fluctuations ``gradual'' with a stable mean; instead, there are pronounced oscillations and spikes up to well above 10, with changing baseline levels.

Option D: Clearly incorrect. The baseline is nowhere near 87–94; values range roughly from 0 to about 23, with no step-like transitions between high plateaus.

The correct answer is B.} \\
\bottomrule
\end{tabularx}
\caption{``Global Pattern Differentiation'' OOD example with \textsc{LLaTiSA}'s response. The question is from MMTS-Bench \cite{mmtsbench}.}
\label{tab:llatisa_l2_global}
\end{table*}

\end{document}